\pdfoutput=1
\documentclass{article}
\usepackage{iclr2027_conference,times}
\usepackage[T1]{fontenc}
\usepackage{amsmath,amssymb,amsfonts,bm}
\usepackage{booktabs,tabularx,array,multirow}
\usepackage{graphicx,xcolor}
\usepackage{wrapfig,needspace}
\usepackage{enumitem}
\usepackage{algorithm,algorithmic}
\usepackage{tikz}
\usepackage{pgfplots}
\pgfplotsset{compat=1.18}
\usetikzlibrary{arrows.meta,positioning}
\usepackage{hyperref}
\usepackage{url}
\hypersetup{colorlinks=true,linkcolor=blue!55!black,citecolor=blue!55!black,urlcolor=blue!55!black}
\title{Leaky Students: Membership Inference against On-Policy Distillation}
\author{Zhexi Lu\thanks{Equal contribution.} \\ Rensselaer Polytechnic Institute \\ \texttt{luz17@rpi.edu}
\And Mingzhi Zhu\footnotemark[1] \\ Rensselaer Polytechnic Institute \\ \texttt{zhum8@rpi.edu}
\AND Stacy Patterson \\ Rensselaer Polytechnic Institute \\ \texttt{sep@cs.rpi.edu}
\And Lei Yu \\ Rensselaer Polytechnic Institute \\ \texttt{yul9@rpi.edu}}

\newcommand{\method}{Leaky}

\newcommand{\pos}[1]{\left[#1\right]_{+}}

\newcommand{\E}{\mathbb{E}}

\newcommand{\score}{S}
\usepackage{longtable,makecell,placeins}
\newcommand{\opdLong}[6]{%
\par\Needspace{6\baselineskip}\begingroup
\small\setlength{\tabcolsep}{3pt}\renewcommand{\arraystretch}{1.15}%
\setlength{\LTleft}{0pt}\setlength{\LTright}{0pt}\setlength{\LTcapwidth}{\linewidth}%
\begin{longtable}{#3}
\caption{#1}\label{#2}\\\toprule
#4
\endfirsthead
\multicolumn{#6}{@{}l@{}}{Table \thetable{} (continued)}\\\toprule
#4
\endhead
\midrule\multicolumn{#6}{@{}r@{}}{Continued on next page}\\\endfoot
\bottomrule\endlastfoot
#5
\end{longtable}\endgroup}
\begin{document}
\maketitle
\begin{abstract}
On-policy distillation (OPD) trains a student to match a teacher's next-token distributions on student-generated trajectories. However, privileged information supplied to the teacher for OPD training may contain sensitive data. Whether the student leaks private information about the records supplied to the teacher during distillation remains poorly understood.
To the best of our knowledge, we present the first systematic study of membership inference in this setting. We find that fresh student trajectories expose sparse membership signals that fixed reference-answer losses often miss. These signals are mixed with probability changes caused by training on other records. We introduce \method{}, which samples fresh trajectories from the target model and compares its token log-probabilities with the maximum across matched reference models trained without the candidate records. It applies Leaky ReLU to the resulting gaps, preserving positive gaps and downweighting negative gaps as an approximate correction for incidental positive gaps in non-members. Across fifteen targets spanning mathematics, medical question answering, and code generation, \method{} outperforms all evaluated baselines and achieves mean AUROC 0.875, compared with 0.614 for the strongest baseline on each target in the main evaluation. On the same sampled trajectories, the strongest baseline achieves mean AUROC 0.826. These results show that students trained through OPD can expose the membership of records used for teacher supervision, even when fixed reference-answer losses provide little evidence of membership.

\end{abstract}

\section{Introduction}\label{sec:introduction}

Reinforcement learning with verifiable rewards (RLVR) has become a key paradigm for improving the reasoning capabilities of large language models (LLMs)~\citep{lambert2024tulu3,deepseek2025r1}. However, RLVR typically relies on sparse, outcome-based rewards that provide limited token-level feedback. OPD addresses this limitation by providing dense, token-level teacher supervision on trajectories sampled from the student's own policy, mitigating the train--inference distribution mismatch of fixed-sequence distillation~\citep{lu2025onpolicydistillation,agarwal2024gkd}.
 Recent methods additionally give the teacher \emph{privileged information}, such as reference answers or expert demonstrations, that is absent from the student's input~\citep{zhao2026opsd,shenfeld2026sdft,ye2026opcd}. The student learns from the teacher's distributions on its sampled trajectories without directly training to reproduce the privileged reference.

When these references contain sensitive information, excluding them from the student's input does not remove their influence on the released model. For example, a teacher might use confidential clinical annotations to supervise a student that sees only the case description. Those annotations shape the teacher's token-level feedback and, consequently, the student's parameter updates. If the resulting changes retain information specific to a training record, the student may reveal that the record was used, even without reproducing its annotation. An attacker who already knows a case and its annotation could thereby learn whether that case contributed to an institution's training collection. This is a membership privacy risk: the sensitive fact is the record's inclusion in training, rather than necessarily its contents~\citep{shokri2017membership}.

Prior work shows that distillation can leak membership information about both teacher and student training data~\citep{jagielski2023students,zhang2025distillationprivacy}. Membership inference attacks (MIAs) on language models use selected token scores~\citep{zhang2025minkpp}, reference-model calibration~\citep{tao2025informia}, and aggregation over text windows~\citep{chen2026wbc}. Related work also audits RLVR through changes in sampled generations~\citep{liu2025diba}. These findings motivate examining membership leakage in the reference-conditioned OPD setting, where a record's reference answer guides teacher feedback on the student's own trajectories.

OPD poses a particular challenge for this evaluation. Unlike supervised fine-tuning (SFT) and fixed-sequence knowledge distillation (KD), OPD matches the teacher's next-token distributions on student-generated trajectories rather than fixed target texts. Even for the same training record, trajectories can vary across iterations, so the record's influence may span different prefixes and continuations. Because OPD does not directly maximize the fixed reference answer's likelihood, that likelihood may miss membership signals in the student's generation behavior. Conventional loss-based MIAs may therefore provide limited separation between members and non-members, as illustrated in Figure~\ref{fig:residue-likelihood}(a). This motivates our central question: \textbf{\textit{Can fresh student trajectories reveal which records were used for OPD training?}}

\begin{figure}[!t]\centering
\includegraphics[width=\linewidth]{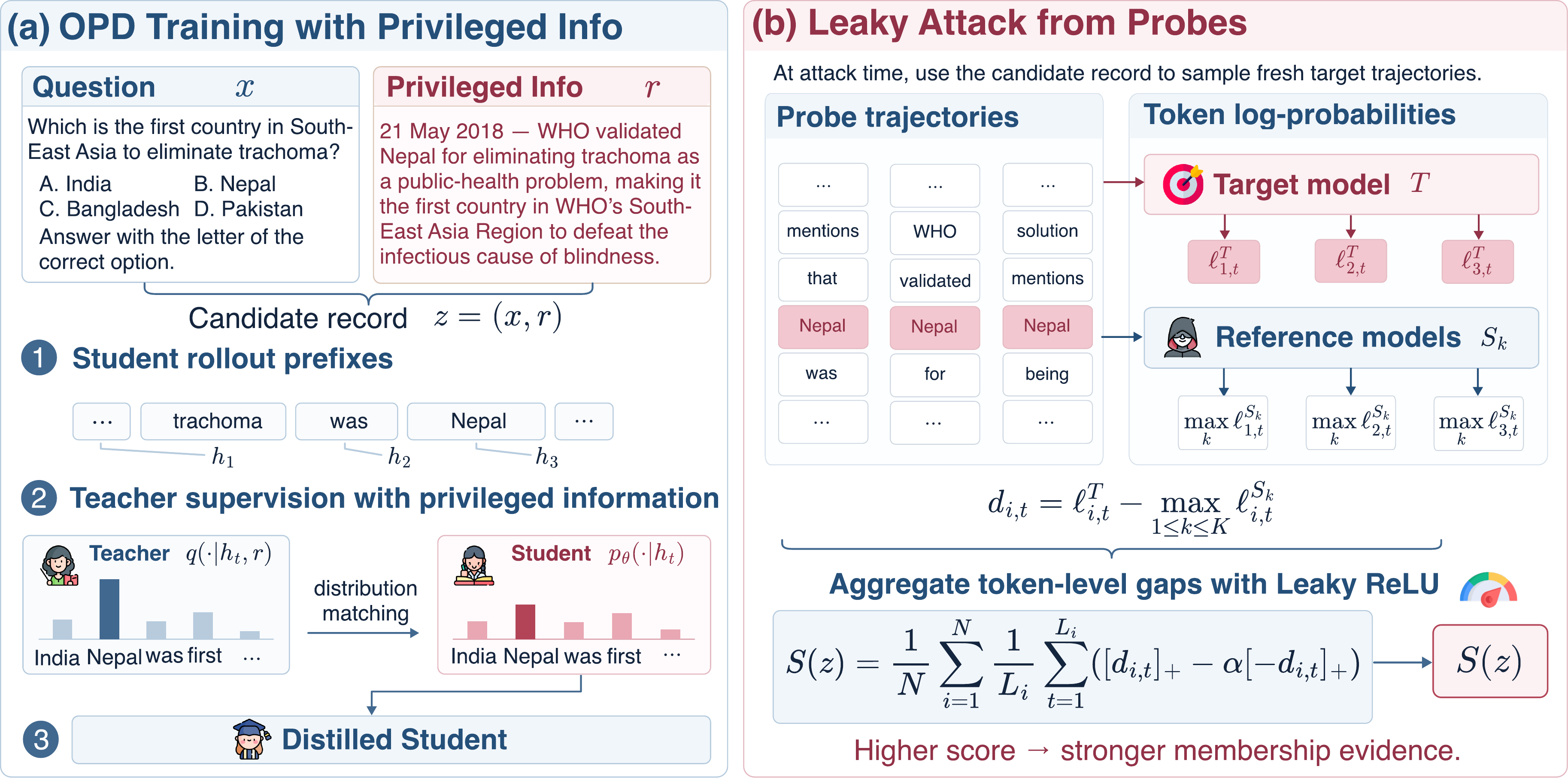}
\caption{Overview. (a) In OPD with privileged information, the student samples rollouts from the prompt $x$ alone, and the teacher, which also sees the reference answer $r$, supplies the next-token distributions the student matches. (b) \method{} samples trajectories from the target conditioned on the candidate records, scores each token under the target and $K$ reference models trained without the candidate, and aggregates the resulting gaps with a Leaky ReLU into the membership score .}\label{fig:residue-pipeline}
\end{figure}
% $d_{i,t}$ (Eq.~\ref{eq:residue-gap})  $\score(z)$ (Eq.~\ref{eq:residue-score})%

We address this question with the following contributions:
\begin{itemize}[leftmargin=*,topsep=2pt,itemsep=1pt,parsep=0pt]
\item \textbf{The First study of membership inference in OPD.} To the best of our knowledge, we present the first systematic study of membership inference on OPD training records (Figure~\ref{fig:residue-pipeline}(a)). We show that fresh student trajectories reveal membership information through local token-probability differences between the target and matched reference models, which fixed reference-answer losses can miss.

\item \textbf{\method, a membership inference attack tailored to OPD.} We introduce \method{} (Figure~\ref{fig:residue-pipeline}(b)), a reference-based attack that infers membership from fresh trajectories sampled from the target model for each candidate record. Instead of scoring the record's fixed reference answer, \method{} calibrates token-level log-probabilities on these trajectories against matched reference models and aggregates the resulting gaps asymmetrically to extract membership signals. The attack requires no access to the trajectories used during training.

\item \textbf{Evaluation across three domains.} We evaluate \method{} on 15 target models spanning mathematics, medical question answering, and code generation. \method{} outperforms all evaluated baselines on every target, improving mean AUROC from 0.614 for the strongest baseline on each target to 0.875. We further demonstrate that membership leakage persists even under text-only prefix queries, show that synthetic auxiliary data can support effective attacks, and evaluate the privacy--utility trade-offs of LoRA and DP-LoRA defenses.
\end{itemize}

\section{Preliminaries}\label{sec:preliminaries}
\subsection{On-Policy Distillation}\label{sec:opd-background}
Knowledge distillation trains a student to match a teacher's output distribution~\citep{hinton2015distilling}. In on-policy distillation (OPD), the student samples a trajectory $\hat y\sim p_\theta(\cdot\mid x)$ for a prompt $x$ and learns from the teacher's next-token distributions on the sampled trajectory~\citep{agarwal2024gkd,lu2025onpolicydistillation}. As the student policy changes during training, subsequent trajectories are sampled from the updated policy. Recent work provides the teacher with privileged information that is absent from the student's input during training~\citep{penaloza2026privileged,zhao2026opsd,shenfeld2026sdft}. We focus on records $z=(x,r)$, where $x$ is a prompt and $r$ is a reference answer used as privileged information. The student samples trajectories conditioned on $x$, while the teacher additionally receives $r$. At each prefix $\hat y_{<t}$, the teacher provides $q(\cdot\mid x,r,\hat y_{<t})$ as supervision for the student's distribution $p_\theta(\cdot\mid x,\hat y_{<t})$. For example, forward-KL distillation uses the token-level objective
\begin{equation}
\mathcal L_t=D_{\mathrm{KL}}\!\left(q(\cdot\mid x,r,\hat y_{<t})\,\middle\|\,p_\theta(\cdot\mid x,\hat y_{<t})\right).\label{eq:opd}
\end{equation}
The reference answer influences training through the teacher's distribution; the student is not directly trained to maximize $\log p_\theta(r\mid x)$.

\subsection{Threat Model}\label{sec:threat-model}
Given a target model $T$ and a candidate record $z=(x,r)$, the attacker aims to determine whether~$z$ belongs to the OPD training set $\mathcal D$ used to train $T$. Both the prompt $x$ and the reference answer $r$ are available to the attacker. In the main gray-box setting, the attacker can sample from the target and query the conditional log-probability of a specified token given a prompt and trajectory prefix. The target's parameters, gradients, and training trajectories are unavailable to the attacker. The attacker has access to the pretrained model used to initialize the target, the teacher model, and the OPD training configuration. We also assume access to auxiliary data drawn from the same task distribution, with the evaluated membership candidates excluded. We consider two additional settings: text-only black-box access without token probabilities, and limited auxiliary data. These settings are evaluated separately in Section~\ref{sec:experiments}.

\section{\method: Membership Inference against OPD}\label{sec:method}

\method{} is a reference-model-based membership inference attack against models trained with OPD. The attacker first trains $K$ reference models on auxiliary data drawn from the same distribution as the training data. For each candidate record, the attacker then samples trajectories from the target with the prompt and privileged information, compares the target's token log-probabilities with the highest log-probability across all the reference models at each position, and uses the resulting gaps aggregated with Leaky ReLU as the membership score. We first motivate our design and then describe the attack procedure in this section.

\subsection{Understanding Membership Signals in OPD}\label{sec:residue-evidence}

\begin{figure}[!t]\centering
\includegraphics[width=\linewidth]{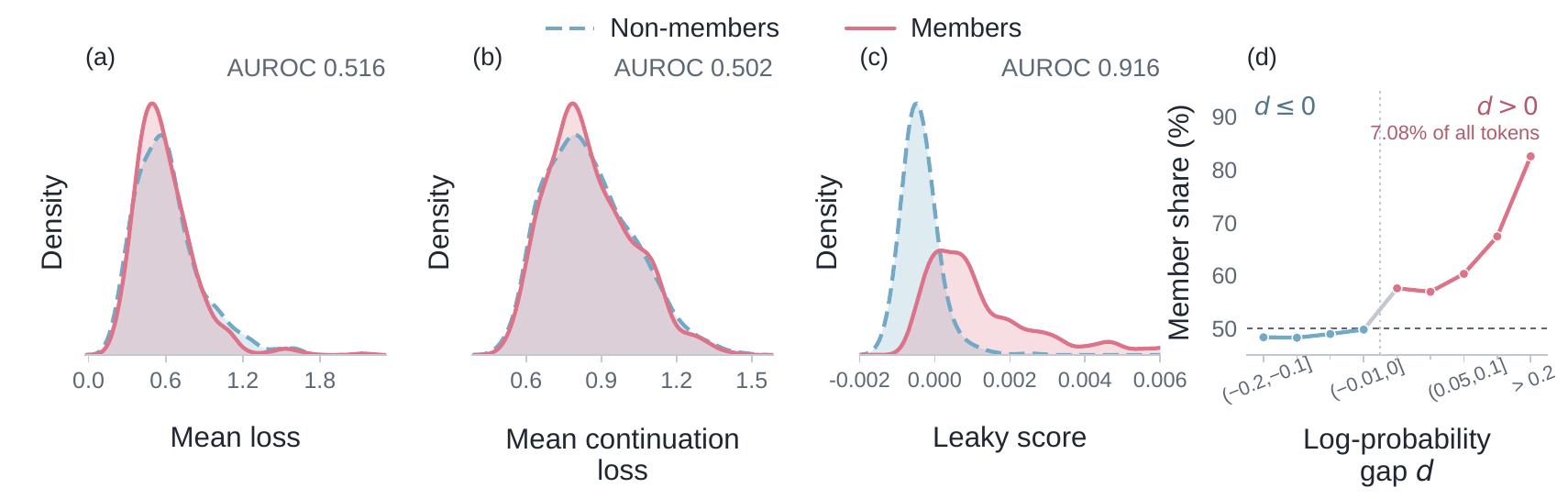}
\caption{Membership signals in OPD. (a,b) Mean loss on fixed answers and sampled continuations provides limited separation between members and non-members. (c) \method{} separates the two groups more clearly. (d) Positive token-level gaps account for only 7.08\% of all tokens, and larger positive gaps generally have a higher member share.}\label{fig:residue-likelihood}
\end{figure}

Loss-based membership inference~\citep{yeom2018privacy} scores a candidate record $(x,r)$ by the target model's loss on the reference answer $r$ given the prompt $x$. SFT minimizes exactly this loss, so the loss can help distinguish members from non-members. OPD optimizes a different objective: the student matches the teacher's next-token distributions by minimizing the per-token distillation loss $\mathcal L_t$ (Eq.~\ref{eq:opd}) on trajectories sampled from its own policy. Training therefore shapes how the student generates without directly raising the likelihood of the fixed reference answer. As a result, a record can leave a trace in the student's generation behavior while its reference-answer loss barely changes. On the Qwen3-1.7B mathematics target in Table~\ref{tab:opd_main}, trained on OpenThoughts-Math, member and non-member losses largely overlap; the Loss attack achieves an AUROC of only 0.516 (Figure~\ref{fig:residue-likelihood}(a)).

We further examine whether membership signals exist on sampled trajectories. For each candidate record $(x,r)$, we sample eight trajectories from the target model conditioned on $(x,r)$ and average the token loss first within each trajectory and then across the eight trajectories. The resulting signals largely overlap between members and non-members, yielding an AUROC of 0.502 (Figure~\ref{fig:residue-likelihood}(b)). To understand what the average loss may miss, we examine both the source of token-probability increases and their distribution across token positions. First, the average loss reflects training on \emph{all} records, not only the candidate: other records of the same task can produce comparable probability gains, so a low loss alone does not indicate that the candidate was used for training. Extracting a membership signal therefore requires \emph{calibrating} the target's token probabilities against models trained without the candidate, so that shared changes are accounted for. Second, stochastic sampling can produce different trajectories for the same $(x,r)$, and training on a member record does not necessarily increase the probability of every sampled token. Averaging over all positions may therefore hide probability changes concentrated at particular positions. Prior work also shows that aggregating selected token-level scores can be more effective than averaging the loss over all tokens~\citep{shi2024mink,zhang2025minkpp}. We examine (1)~how to calibrate token probabilities using reference models and (2)~which token positions carry membership signals and how to aggregate their scores.

\noindent\textbf{(1) Shared changes in token probabilities.} To separate the effect of the candidate from these shared changes, we need to know what training without the candidate already produces. The pretrained base model cannot provide this: its difference from the target reflects the effect of training on every record, not only the candidate. Following~\citet{zarifzadeh2024rmia} and~\citet{tao2025informia}, we train $K$ reference models on auxiliary data that exclude the candidate records. At each position of a sampled trajectory, the target and the reference models score the same token under the same prompt and preceding tokens. A high target probability thus counts as membership signal only when it exceeds the probabilities of all reference models at that position. The challenge is that such positions are sparse and can also arise for non-members; we examine this empirically next.

\noindent\textbf{(2) Membership signal in token-level gaps.} We define the gap $d$ as the target's token log-probability minus the highest reference log-probability. A positive gap means that the target exceeds every reference model at that position. In the same mathematics experiment, only 7.08\% of scored tokens have positive gaps. Within the positive-gap region, the member share generally increases with the gap and reaches approximately 83\% in the highest-gap interval. In contrast, the member share among negative-gap positions is 49.3\% (Figure~\ref{fig:residue-likelihood}(d)). These observations motivate treating positive gaps as the primary membership signal. Non-members also produce positive gaps, however, so a positive gap cannot be attributed entirely to training on the candidate. Greater variation in token predictions across models can produce both larger positive and larger negative gaps, even for non-members, which suggests using the aggregated negative contribution as an approximate correction for such variation. Most scored tokens have negative gaps, and the magnitude of this contribution varies across records. Giving negative gaps equal weight may obscure membership differences carried by the positive gaps. We therefore retain positive gaps at full weight and assign negative gaps a small weight through Leaky ReLU, defined in Section~\ref{sec:residue-attack}. Further details are in Appendix~\ref{app:leaky-negative-correction}.

\subsection{Attack Method}\label{sec:residue-attack}
\method{} consists of three steps: (1) trajectory sampling from the target, (2) reference-model calibration of token log-probabilities to account for shared changes, and (3) Leaky ReLU aggregation of the resulting gaps.

\noindent\textbf{Step 1: Trajectory sampling.} Given a candidate record $z=(x,r)$, we sample $N$ trajectories $y_1,\ldots,y_N$ from the target model $T$ conditioned on $(x,r)$. Sampling trajectories from the target allows us to examine its generation behavior after OPD. The attack samples these trajectories at attack time; it does not need the trajectories sampled during training.

\noindent\textbf{Step 2: Reference-model calibration.} The goal of this step is to identify token positions where the target's probability is higher than what models trained without the candidate can achieve. For each sampled token $y_{i,t}$, we compute the conditional log-probability under the target $T$ and under each reference model $S_k$:
\begin{equation}
\ell^M_{i,t}=\log p_M(y_{i,t}\mid x,\,y_{i,<t}),\qquad M\in\{T,S_1,\ldots,S_K\}.
\label{eq:residue-logprob}
\end{equation}
All models are conditioned on the prompt $x$ and the trajectory prefix $y_{i,<t}$. The reference answer $r$ is used during trajectory sampling (Step~1) but is not included in the scoring input. Because training on other records can raise the same token probabilities that training on the candidate raises, we compare the target against reference models trained without the candidate~\citep{zarifzadeh2024rmia,tao2025informia}. At each position, we compute the gap between the target's log-probability and the \emph{highest} reference log-probability:
\begin{equation}
d_{i,t}=\ell^T_{i,t}-\max_{1\leq k\leq K}\ell^{S_k}_{i,t}.\label{eq:residue-gap}
\end{equation}
We take the maximum rather than the mean because a target probability above the mean may still fall within the range of the reference models. A positive $d_{i,t}$ means that the target exceeds \emph{every} reference model at that position, providing a stricter membership signal; Figure~\ref{fig:residue-likelihood}(d) shows a higher member share for positive gaps. The maximum is taken independently at each position, so the reference model that attains it can differ across positions.

\noindent\textbf{Step 3: Leaky ReLU aggregation.} Let $L_i$ be the number of scored tokens in the $i$-th trajectory. The membership score is
\begin{equation}
\score(z)=\frac{1}{N}\sum_{i=1}^{N}\frac{1}{L_i}\sum_{t=1}^{L_i}\left(\pos{d_{i,t}}-\alpha\pos{-d_{i,t}}\right),\label{eq:residue-score}
\end{equation}
where $\pos{a}=\max\{a,0\}$ denotes the positive part. The transformation $\pos{a}-\alpha\pos{-a}$ is the leaky rectified linear unit (Leaky ReLU)~\citep{maas2013rectifier}, which reduces to ReLU~\citep{nair2010relu,glorot2011rectifier} when $\alpha=0$. We apply Leaky ReLU to each token-level gap before averaging. Positive gaps are retained unchanged, whereas negative gaps are scaled by the small coefficient $\alpha$. We include the downweighted negative component as an approximate correction to the positive contribution. We first average over all $L_i$ scored tokens within each trajectory and then average the $N$ trajectory scores with equal weight. A higher $\score(z)$ indicates stronger membership signal. Figure~\ref{fig:residue-likelihood}(c) shows the distributions of $\score(z)$ for members and non-members. The full procedure is summarized in Algorithm~\ref{alg:residue}.

\begin{algorithm}[t]
\caption{\method: membership inference using sampled trajectories}\label{alg:residue}
\begin{algorithmic}[1]
\REQUIRE Candidate record $z=(x,r)$; target $T$; reference models $S_1,\ldots,S_K$ trained without $z$; number of trajectories $N$; negative slope $\alpha$
\ENSURE Membership score $\score(z)$
\STATE Sample $y_1,\ldots,y_N$ from $T$, conditioned on $(x,r)$.
\FOR{$i=1,\ldots,N$}
\STATE Let $L_i$ be the number of scored tokens in $y_i$.
\FOR{$t=1,\ldots,L_i$}
\STATE Compute $\ell^M_{i,t}=\log p_M(y_{i,t}\mid x,y_{i,<t})$ for $M\in\{T,S_1,\ldots,S_K\}$.
\STATE $d_{i,t}\leftarrow\ell^T_{i,t}-\max_k\ell^{S_k}_{i,t}$
\ENDFOR
\STATE $s_i\leftarrow \frac{1}{L_i}\sum_{t=1}^{L_i}\left(\pos{d_{i,t}}-\alpha\pos{-d_{i,t}}\right)$
\ENDFOR
\RETURN $\frac{1}{N}\sum_{i=1}^N s_i$
\end{algorithmic}\end{algorithm}

\section{Experiments}
\label{sec:experiments}\label{sec:opd_mia_experiments}
We evaluate \method{} against OPD-trained models across mathematics, medical question answering, and code generation, using five models from three model families. We compare it with existing MIAs and report downstream task utility alongside attack performance. We further examine black-box access and evaluate LoRA~\citep{hu2021lora} and DP-LoRA~\citep{yu2022dpfinetuning} as potential defenses. Finally, we study the effects of reference-model count and trajectory
generation, assess membership leakage under on-policy self-distillation (OPSD)~\citep{zhao2026opsd}, and evaluate the use of synthetic auxiliary data for reference-model training.

\subsection{Experimental Setup}\label{sec:opd_mia_setup}

\noindent\textbf{Models and datasets.} We evaluate five students (Qwen3-1.7/4B~\citep{qwen2025qwen3}, Qwen3.5-2/4B~\citep{qwen2026qwen35}, and Gemma-4-E2B~\citep{gemma2026gemma4}) across three model families on OpenThoughts-Math~\citep{openthoughts2025}, MedMCQA~\citep{pal2022medmcqa}, and TACO~\citep{li2023taco}, covering mathematics, medical question answering, and code generation. Each candidate pool contains 1,024 records: 512 members and 512 non-members. All candidates are excluded from reference-model training. We describe the data splits in Appendix~\ref{app:opd_mia_cohorts}, and the details of the training configs are listed in Appendix~\ref{app:opd_mia_training}

\noindent\textbf{Metrics.} We report AUROC~\citep{fawcett2006roc} and TPR at 1\%FPR~\citep{carlini2022lira} in the main tables, with 5\% and 10\% FPR results in Appendix~\ref{app:opd_mia_results}. Downstream utility is Avg@8 on AIME 2024~\citep{maa2024aime}, MedQA-USMLE~\citep{jin2021medqa}, and LiveCodeBench v6~\citep{jain2025livecodebench} at the attacked checkpoint, with identical evaluation settings for the base and trained models. Scoring and evaluation details are in Appendices~\ref{app:opd_mia_primary_protocol} and~\ref{app:opd_mia_utility}.

\noindent\textbf{Baselines.}
We compare \method{} with reference-free attacks, including LOSS~\citep{yeom2018privacy}, ZLIB~\citep{carlini2021extracting}, Min-K\%~\citep{shi2024mink}, Min-K\%++~\citep{zhang2025minkpp}, and ICP-MIA~\citep{lu2026icpmia}, and reference-based attacks, including Ratio~\citep{watson2022difficulty}, WBC~\citep{chen2026wbc}, EZ-MIA~\citep{ilic2026ezmia}, LiRA~\citep{carlini2022lira}, RMIA~\citep{zarifzadeh2024rmia}, InfoRMIA~\citep{tao2025informia}, and DIBA~\citep{liu2025diba}. Ratio, WBC, EZ-MIA, and DIBA use the base model, while LiRA, RMIA, and InfoRMIA use the same eight trained reference models as \method. All methods are evaluated on the same candidate records. Implementation details and hyperparameters are provided in Appendix~\ref{app:opd_mia_baselines}.

\subsection{Main Results}\label{sec:opd_mia_main}\label{sec:opd_mia_baselines}

\begin{table}[t]
\centering
\fontsize{8}{9.2}\selectfont\setlength{\tabcolsep}{2pt}\renewcommand{\arraystretch}{1.0}
\caption{Membership inference against OPD: AUROC and TPR @ 1\% FPR, leading zeros omitted, best per column in bold. \method{} scores eight target-generated trajectories per candidate. Q3/Q3.5: Qwen3/Qwen3.5; G4: Gemma-4-E2B. Results at 5\% and 10\% FPR are in Appendix~\ref{app:opd_mia_results}.}\label{tab:opd_main}
\begin{tabular*}{\linewidth}{@{\extracolsep{\fill}}l*{15}{c}@{}}
\toprule
 & \multicolumn{5}{c}{Mathematics} & \multicolumn{5}{c}{Medical QA} & \multicolumn{5}{c}{Code generation}\\
\cmidrule(lr){2-6}\cmidrule(lr){7-11}\cmidrule(lr){12-16}
Method & \makecell{Q3\\1.7B} & \makecell{Q3\\4B} & \makecell{Q3.5\\2B} & \makecell{Q3.5\\4B} & \makecell{G4\\E2B} & \makecell{Q3\\1.7B} & \makecell{Q3\\4B} & \makecell{Q3.5\\2B} & \makecell{Q3.5\\4B} & \makecell{G4\\E2B} & \makecell{Q3\\1.7B} & \makecell{Q3\\4B} & \makecell{Q3.5\\2B} & \makecell{Q3.5\\4B} & \makecell{G4\\E2B}\\
\midrule
\multicolumn{16}{c}{\textit{AUROC}}\\
\midrule
LOSS & .516 & .513 & .502 & .502 & .505 & .553 & .552 & .542 & .545 & .530 & .546 & .532 & .543 & .550 & .559\\
ZLIB & .528 & .525 & .497 & .497 & .503 & .509 & .511 & .503 & .505 & .504 & .616 & .590 & .612 & .616 & .622\\
Min-K\% & .521 & .518 & .499 & .500 & .506 & .540 & .542 & .532 & .544 & .518 & .535 & .512 & .533 & .539 & .540\\
Min-K\%++ & .521 & .522 & .509 & .509 & .515 & .492 & .508 & .511 & .515 & .522 & .492 & .454 & .502 & .526 & .442\\
ICP-MIA & .522 & .542 & .530 & .518 & .511 & .539 & .543 & .530 & .534 & .524 & .552 & .510 & .529 & .544 & .516\\
Ratio & .502 & .497 & .473 & .504 & .497 & .486 & .469 & .519 & .513 & .470 & .470 & .489 & .516 & .552 & .529\\
WBC & .497 & .504 & .477 & .501 & .517 & .499 & .476 & .500 & .495 & .479 & .469 & .459 & .569 & .618 & .447\\
EZ-MIA & .506 & .517 & .469 & .496 & .493 & .534 & .492 & .482 & .487 & .481 & .466 & .489 & .522 & .556 & .543\\
DIBA & .513 & .489 & .495 & .488 & .472 & .623 & .599 & .589 & .518 & .565 & .544 & .535 & .484 & .570 & .556\\
\midrule
LiRA & .661 & .660 & .534 & .564 & .497 & .712 & .622 & .637 & .573 & .478 & .496 & .562 & .577 & .574 & .506\\
RMIA & .662 & .670 & .518 & .520 & .488 & .698 & .589 & .630 & .570 & .480 & .512 & .563 & .526 & .572 & .509\\
InfoRMIA & .666 & .663 & .561 & .573 & .517 & .740 & .649 & .613 & .579 & .496 & .507 & .558 & .577 & .583 & .527\\
\midrule
\textbf{\method{}} & \textbf{.916} & \textbf{.922} & \textbf{.913} & \textbf{.925} & \textbf{.832} & \textbf{.966} & \textbf{.848} & \textbf{.868} & \textbf{.837} & \textbf{.826} & \textbf{.847} & \textbf{.843} & \textbf{.901} & \textbf{.846} & \textbf{.839}\\
\midrule
\multicolumn{16}{c}{\textit{TPR at 1\% FPR}}\\
\midrule
LOSS & .006 & .004 & .008 & .014 & .006 & .018 & .018 & .016 & .006 & .012 & .006 & .002 & .006 & .008 & .004\\
ZLIB & .004 & .006 & .021 & .023 & .018 & .014 & .012 & .010 & .010 & .006 & .004 & .010 & .002 & .002 & .008\\
Min-K\% & .006 & .004 & .008 & .014 & .006 & .002 & .010 & .004 & .004 & .004 & .006 & .002 & .006 & .010 & .004\\
Min-K\%++ & .000 & .002 & .006 & .010 & .006 & .012 & .014 & .010 & .018 & .012 & .002 & .004 & .004 & .004 & .018\\
ICP-MIA & .006 & .016 & .010 & .004 & .006 & .006 & .014 & .010 & .016 & .008 & .016 & .002 & .002 & .012 & .006\\
Ratio & .010 & .012 & .010 & .012 & .004 & .006 & .010 & .016 & .010 & .004 & .012 & .004 & .004 & .021 & .021\\
WBC & .010 & .012 & .002 & .006 & .014 & .016 & .010 & .023 & .027 & .004 & .008 & .008 & .014 & .061 & .014\\
EZ-MIA & .004 & .006 & .014 & .014 & .010 & .008 & .010 & .020 & .010 & .004 & .010 & .006 & .037 & .021 & .000\\
DIBA & .014 & .020 & .006 & .010 & .012 & .016 & .039 & .018 & .020 & .006 & .018 & .016 & .012 & .014 & .012\\
\midrule
LiRA & .066 & .055 & .035 & .014 & .008 & .148 & .045 & .023 & .021 & .018 & .012 & .035 & .021 & .020 & .018\\
RMIA & .084 & .068 & .041 & .025 & .010 & .164 & .029 & .041 & .076 & .008 & .016 & .035 & .016 & .027 & .008\\
InfoRMIA & .064 & .070 & .039 & .016 & .023 & .150 & .070 & .053 & .025 & .000 & .014 & .035 & .033 & .027 & .021\\
\midrule
\textbf{\method{}} & \textbf{.418} & \textbf{.576} & \textbf{.477} & \textbf{.582} & \textbf{.207} & \textbf{.740} & \textbf{.449} & \textbf{.512} & \textbf{.461} & \textbf{.350} & \textbf{.434} & \textbf{.396} & \textbf{.578} & \textbf{.527} & \textbf{.270}\\
\bottomrule
\end{tabular*}
\end{table}

\noindent\textbf{Attack performance.}
Table~\ref{tab:opd_main} compares \method{} with the evaluated baselines across mathematics, medical question answering, and code generation. \method{} achieves the highest AUROC and TPR at 1\% FPR on all fifteen targets, demonstrating its effectiveness both overall and at a low false-positive rate. Averaged over these targets, \method{} improves AUROC from 0.614 for the strongest baseline on each target to 0.875. Several baselines improve substantially when applied to the same sampled trajectories used by \method{} (Table~\ref{tab:opd_cont_baselines}), further supporting our motivation to examine membership signals in the student's generation behavior rather than relying on fixed reference-answer losses. Under this comparison, \method{} still outperforms all evaluated baselines on every target, supporting the effectiveness of its token-level reference-model calibration and leaky ReLU aggregation. All attacked models also outperform their base models on the corresponding downstream benchmarks (Table~\ref{tab:opd_utility}). These results show that our OPD training can indeed improve downstream task performance while being vulnerable to membership inference attacks.

\noindent\textbf{Black-box attack.}\label{sec:opd_mia_blackbox} \method{} requires token log-probabilities, which a text-only interface may not provide. We adapt it to this setting by querying the target with trajectory prefixes and recording whether its greedy prediction matches the sampled token. These binary indicators replace token log-probabilities in the gap computation and leaky ReLU aggregation; Appendix~\ref{app:opd_mia_blackbox} provides the scoring formula. As shown in Table~\ref{tab:opd_mia_blackbox}, the text-only attack remains effective across all fifteen targets, including at low false-positive rates. Restricting access to token probabilities alone therefore does not prevent membership leakage in OPD.

\begin{table}[t]
\centering
\fontsize{8}{9.2}\selectfont\setlength{\tabcolsep}{3pt}\renewcommand{\arraystretch}{1.0}
\caption{Black-box membership inference on the fifteen targets in Table~\ref{tab:opd_main}. We report AUROC and TPR at 1\% FPR for the text-only prefix-query attack.}\label{tab:opd_mia_blackbox}
\begin{tabular*}{\linewidth}{@{\extracolsep{\fill}}l*{6}{c}@{}}
\toprule
 & \multicolumn{2}{c}{Mathematics} & \multicolumn{2}{c}{Medical QA} & \multicolumn{2}{c}{Code generation}\\
\cmidrule(lr){2-3}\cmidrule(lr){4-5}\cmidrule(lr){6-7}
Target model & AUROC & TPR@1\% FPR & AUROC & TPR@1\% FPR & AUROC & TPR@1\% FPR\\
\midrule
Qwen3-1.7B & .776 & .273 & .854 & .389 & .747 & .213\\
Qwen3-4B & .796 & .289 & .691 & .211 & .687 & .146\\
Qwen3.5-2B & .774 & .191 & .768 & .342 & .740 & .238\\
Qwen3.5-4B & .812 & .309 & .737 & .189 & .788 & .412\\
Gemma-4-E2B & .727 & .152 & .698 & .189 & .660 & .084\\
\bottomrule
\end{tabular*}
\end{table}

\subsection{Potential Defenses}\label{sec:opd_mia_defenses}
We evaluate LoRA~\citep{hu2021lora} and DP-LoRA~\citep{yu2022dpfinetuning} as potential defenses on
Qwen3-1.7B with a Qwen3-8B teacher across the three domains. DP-LoRA applies record-level gradient clipping and Gaussian noise~\citep{abadi2016dp} at a nominal privacy budget of $\varepsilon\approx8$.
Training data and evaluation candidates are held fixed within each domain, and each configuration uses eight reference models trained with the corresponding procedure. Training settings, privacy accounting, and additional results are provided in Appendix~\ref{app:opd_mia_defense}.

\begin{table}[!htbp]
\centering
\fontsize{8}{9.2}\selectfont\setlength{\tabcolsep}{3pt}\renewcommand{\arraystretch}{1.0}
\caption{Defenses on Qwen3-1.7B (teacher Qwen3-8B): change in Avg@8 over the base (percentage points), AUROC, and TPR at 1\% FPR of \method{} with eight matched reference models. Full fine-tuning is the Table~\ref{tab:opd_main} cell; DP-LoRA is at nominal $\varepsilon\approx8$. Configurations are in Appendix~\ref{app:opd_mia_defense}.}\label{tab:opd_mia_defense}
\begin{tabular*}{\linewidth}{@{\extracolsep{\fill}}lrccrccrcc@{}}
\toprule
 & \multicolumn{3}{c}{Mathematics} & \multicolumn{3}{c}{Medical QA} & \multicolumn{3}{c}{Coding}\\
\cmidrule(lr){2-4}\cmidrule(lr){5-7}\cmidrule(lr){8-10}
Training & $\Delta$Util. & AUROC & TPR & $\Delta$Util. & AUROC & TPR & $\Delta$Util. & AUROC & TPR\\\midrule
Full fine-tuning & $+5.00$ & 0.9158 & 0.418 & $+3.98$ & 0.9665 & 0.740 & $+0.57$ & 0.8467 & 0.434\\
LoRA & $+2.50$ & 0.6455 & 0.098 & $+2.46$ & 0.7868 & 0.268 & $+0.14$ & 0.6585 & 0.146\\
DP-LoRA ($\varepsilon\approx8$) & $+1.67$ & 0.4903 & 0.014 & $+0.04$ & 0.5044 & 0.023 & $+0.07$ & 0.5097 & 0.027\\
\bottomrule
\end{tabular*}
\end{table}

\noindent\textbf{Defense performance.} Table~\ref{tab:opd_mia_defense} reports attack performance alongside changes in downstream utility relative to the base model.
LoRA reduces both AUROC and TPR at 1\% FPR compared with full fine-tuning across all three domains, but \method{} still achieves AUROCs of 0.646--0.787, indicating that membership information
remains detectable. LoRA also retains positive utility gains, although these are smaller than those of full fine-tuning. Adding clipping and noise brings AUROC close to chance and reduces TPR at 1\% FPR to 0.014--0.027, but further limits the downstream gains. DP-LoRA retains a 1.67 percentage-point improvement in AIME Avg@8 over the base model, whereas its gains on medical QA and code
generation are below 0.1 percentage points. These results illustrate a trade-off between resistance to
\method{} and downstream utility in the evaluated configurations.

\subsection{Ablation Study}\label{sec:opd_mia_ablations}
We examine the effects of reference-model count, trajectory generation,
and self-distillation. Additional ablations of reference-model calibration
and token aggregation are provided in
Appendix~\ref{app:opd_mia_aggregation}.
\begin{figure}[t]\centering
% BEGIN INLINE FIGURE: ablation_ab
\definecolor{abBlue}{HTML}{74A9C5}\definecolor{abRose}{HTML}{DD7389}\definecolor{abSand}{HTML}{C4AC5E}%
\pgfplotsset{ablation panel/.style={scale only axis,width=2.05cm,height=2.05cm,axis lines*=left,axis line style={black!40,line width=.4pt},tick style={black!40,line width=.4pt},tick align=outside,major tick length=2pt,tick label style={font=\footnotesize,text=black!70},label style={font=\footnotesize,text=black!85},ylabel shift=-5pt,xlabel style={text depth=.3ex},grid style={black!10,line width=.3pt},scaled ticks=false,clip=false,legend style={font=\scriptsize,draw=none,fill=none,inner sep=1pt,row sep=-2.5pt,cells={anchor=west}},legend image code/.code={\draw[mark repeat=2,mark phase=2,##1] plot coordinates {(0cm,0cm) (.15cm,0cm) (.3cm,0cm)};}},ablation series/.style={color=#1,line width=.9pt,mark=*,mark size=1.6pt,mark options={fill=#1,draw=white,line width=.4pt}}}\tikzset{ablation title/.style={anchor=south,yshift=5pt,font=\small}}%
\begin{tikzpicture}
\begin{axis}[ablation panel,name=a,at={(0,0)},anchor=south west,xlabel={Reference models $K$},ylabel={AUROC},xmin=.6,xmax=8.4,xtick={1,...,8},ymin=.52,ymax=.98,ytick={.6,.7,.8,.9},ymajorgrids,y tick label style={/pgf/number format/fixed,/pgf/number format/precision=1},legend pos=south east]
\addplot[ablation series=abBlue] coordinates {(1,.7312) (2,.7861) (3,.8370) (4,.8579) (5,.8850) (6,.8979) (7,.9070) (8,.9158)};\addlegendentry{Mathematics}
\addplot[ablation series=abRose,mark=square*,mark size=1.5pt] coordinates {(1,.9103) (2,.9432) (3,.9519) (4,.9592) (5,.9633) (6,.9665) (7,.9678) (8,.9665)};\addlegendentry{Medical}
\addplot[ablation series=abSand,mark=triangle*,mark size=2pt] coordinates {(1,.6918) (2,.7550) (3,.7823) (4,.7996) (5,.8201) (6,.8337) (7,.8408) (8,.8467)};\addlegendentry{Coding}
\end{axis}
\begin{axis}[ablation panel,name=b,at={(a.south east)},anchor=south west,xshift=1.15cm,xlabel={Reference models $K$},ylabel={TPR at 1\% FPR},xmin=.6,xmax=8.4,xtick={1,...,8},ymin=0,ymax=.8,ytick={0,.2,.4,.6,.8},ymajorgrids,y tick label style={/pgf/number format/fixed,/pgf/number format/precision=1}]
\addplot[ablation series=abBlue] coordinates {(1,.162) (2,.312) (3,.301) (4,.365) (5,.373) (6,.400) (7,.414) (8,.418)};
\addplot[ablation series=abRose,mark=square*,mark size=1.5pt] coordinates {(1,.475) (2,.578) (3,.609) (4,.689) (5,.676) (6,.672) (7,.666) (8,.740)};
\addplot[ablation series=abSand,mark=triangle*,mark size=2pt] coordinates {(1,.176) (2,.229) (3,.266) (4,.301) (5,.369) (6,.424) (7,.422) (8,.434)};
\end{axis}
\begin{axis}[ablation panel,name=c,at={(b.south east)},anchor=south west,xshift=2.00cm,xbar,bar width=6pt,bar shift=0pt,xlabel={AUROC},xmin=.8,xmax=.95,xtick={.8,.85,.9,.95},x tick label style={font=\footnotesize,/pgf/number format/fixed,/pgf/number format/precision=2},ymin=.5,ymax=6.5,y dir=reverse,ytick={1,2,3,4,5,6},yticklabels={Qwen3-14B,Qwen3-8B,Qwen3-4B,Qwen3-1.7B,{Target w/o $r$},Target},y tick label style={align=right,font=\footnotesize\linespread{.9}\selectfont},ytick style={draw=none},xmajorgrids,clip=true,clip mode=individual,nodes near coords,point meta=rawx,nodes near coords style={anchor=west,xshift=1pt,font=\scriptsize,text=black!70,/pgf/number format/fixed,/pgf/number format/fixed zerofill,/pgf/number format/precision=3}]
\addplot[fill=abBlue,draw=none] coordinates {(.904,1) (.904,2) (.899,3) (.857,4) (.880,5)};
\addplot[fill=abRose,draw=none] coordinates {(.916,6)};
\end{axis}
\begin{axis}[ablation panel,name=d,at={(c.south east)},anchor=south west,xshift=1.20cm,xlabel={Training step},ylabel={AUROC},xmin=17,xmax=108,xtick={25,50,75,100},ymin=.5,ymax=1,ytick={.5,.6,.7,.8,.9,1},ymajorgrids,y tick label style={/pgf/number format/fixed,/pgf/number format/fixed zerofill,/pgf/number format/precision=1}]
\addplot[ablation series=abBlue,forget plot] coordinates {(25,.7259) (50,.8088) (75,.8654) (100,.8771)} node[pos=1,anchor=south,yshift=2pt,font=\footnotesize,text=black!85]{OPD};
\addplot[ablation series=abRose,forget plot] coordinates {(25,.5935) (50,.6565) (75,.6869) (100,.6987)} node[pos=1,anchor=north,yshift=-2pt,font=\footnotesize,text=black!85]{OPSD};
\end{axis}
\node[ablation title] at (a.north) {(a) AUROC vs.\ $K$};
\node[ablation title] at (b.north) {(b) TPR vs.\ $K$};
\node[ablation title] at (c.north) {(c) Sampling model};
\node[ablation title] at (d.north) {(d) Self-distillation};
\end{tikzpicture}%

% END INLINE FIGURE: ablation_ab
\caption{Ablations on Qwen3-1.7B. (a) AUROC and (b) TPR at 1\% FPR against the number of reference models $K$ in the three domains; $K=8$ is Table~\ref{tab:opd_main}, and the other students are in Figures~\ref{fig:opd_shadow_sweep} and~\ref{fig:opd_shadow_tpr}. (c) Mathematics AUROC. (d) Mathematics AUROC across checkpoints for OPD with the Qwen3-8B teacher and for OPSD with the frozen initial student.}\label{fig:opd_ablation}
\end{figure}
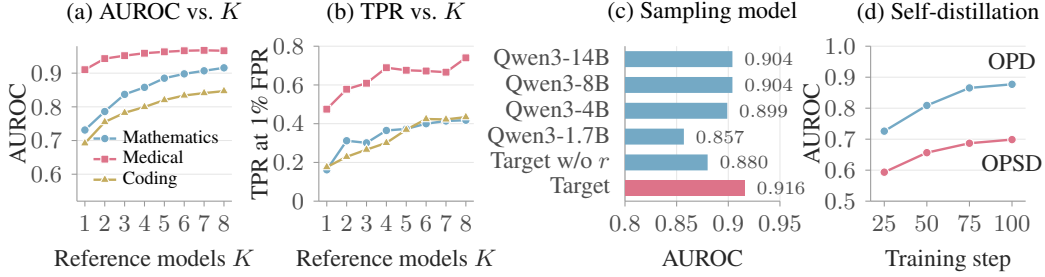

\noindent\textbf{Number of reference models.}
Figures~\ref{fig:opd_ablation}(a,b) show the effect of reference-model count on Qwen3-1.7B across the three domains, with results for the
remaining students in Appendix~\ref{app:opd_mia_shadow_ablation}.
Increasing $K$ from one to eight improves both AUROC and TPR at 1\% FPR on all fifteen targets, with a mean AUROC gain of 0.15. Four reference models already capture most of this gain on fourteen targets, although eight achieve higher AUROC on every target.
These results indicate that a small reference ensemble provides useful calibration, while additional models further improve attack performance.

\noindent\textbf{Trajectory generation.}
Figure~\ref{fig:opd_ablation}(c) compares trajectory-generation choices on the mathematics Qwen3-1.7B target, with the target and reference models held fixed. Trajectories generated by the target, the training teacher, and two alternative models yield similar AUROCs of 0.899--0.916, whereas the frozen base model yields 0.857. Providing the reference answer during target sampling improves AUROC from 0.880 to 0.916. These results support target-generated trajectories as an effective choice and show that reference-answer conditioning strengthens, but is not necessary for, the observed membership signal. Further details are provided in Appendix~\ref{app:opd_mia_generator_pairing}.

\noindent\textbf{OPSD.}
We further evaluate \method{} under OPSD setup, where the teacher is a frozen copy of the initial student conditioned on the problem and reference answer~\citep{zhao2026opsd,shenfeld2026sdft}. On mathematics Qwen3-1.7B, the OPSD run has lower AUROC than the external-teacher OPD run at every evaluated checkpoint, while AUROC increases with training under both procedures (Figure~\ref{fig:opd_ablation}(d)).
These results show that membership leakage also persists in the evaluated self-distilled student. Full settings and results are provided in Appendix~\ref{app:opd_mia_opsd}.

\subsection{Auxiliary Data for Reference Models}\label{sec:opd_mia_imitation}
\begin{table}[t]
\centering
\fontsize{8}{9.2}\selectfont\setlength{\tabcolsep}{3pt}\renewcommand{\arraystretch}{1.0}
\caption{Auxiliary data for reference-model training under the Table~\ref{tab:opd_main} protocol; the real-data and OpenThoughts-Math columns are the Table~\ref{tab:opd_main} cells. (a) Qwen3-1.7B targets, reference models trained on real or Qwen3.5-27B-synthesized records. (b) Mathematics targets, reference models trained on OpenThoughts-Math or DeepMath-103K.}\label{tab:opd_mia_imitation}
\vspace{3pt}
\begin{minipage}[t]{0.47\linewidth}
\centering
\textbf{(a)} Synthetic auxiliary data\par\smallskip
\definecolor{abBlue}{HTML}{74A9C5}\definecolor{abRose}{HTML}{DD7389}%
\pgfplotsset{aux panel/.style={scale only axis,width=2.3cm,height=1.6cm,axis lines*=left,axis line style={black!40,line width=.4pt},tick style={black!40,line width=.4pt},tick align=outside,major tick length=2pt,tick label style={font=\fontsize{6}{7}\selectfont,text=black!70},label style={font=\fontsize{6}{7}\selectfont,text=black!85},ylabel shift=-4pt,grid style={black!10,line width=.3pt},scaled ticks=false,clip=false,ybar,bar width=5pt,xmin=.45,xmax=3.55,xtick={1,2,3},xticklabels={Math,Medical,Code},x tick label style={yshift=1pt},ymajorgrids,nodes near coords,every node near coord/.append style={font=\fontsize{5}{5}\selectfont,text=black!75,yshift=-1.5pt,/pgf/number format/fixed,/pgf/number format/fixed zerofill,/pgf/number format/precision=3,/pgf/number format/skip 0.},legend style={font=\fontsize{6}{6}\selectfont,draw=none,fill=none,inner sep=1pt,column sep=4pt,cells={anchor=west},legend columns=2,at={(1.22,1.12)},anchor=south},legend image code/.code={\fill[##1] (0cm,-.06cm) rectangle (.2cm,.06cm);}}}%
\begin{tikzpicture}
\begin{axis}[aux panel,name=aa,at={(0,0)},anchor=south west,ylabel={AUROC},ymin=.5,ymax=1,ytick={.5,.75,1}]
\addplot[fill=abRose,draw=none,every node near coord/.append style={xshift=-2.6pt}] coordinates {(1,.916) (2,.966) (3,.847)};\addlegendentry{Real data}
\addplot[fill=abBlue,draw=none,every node near coord/.append style={xshift=2.6pt}] coordinates {(1,.802) (2,.727) (3,.616)};\addlegendentry{Synthetic data}
\end{axis}
\begin{axis}[aux panel,name=ab,at={(aa.south east)},anchor=south west,xshift=0.85cm,ylabel={TPR at 1\% FPR},ymin=0,ymax=.8,ytick={0,.4,.8}]
\addplot[fill=abRose,draw=none,every node near coord/.append style={xshift=-2.6pt}] coordinates {(1,.418) (2,.740) (3,.434)};
\addplot[fill=abBlue,draw=none,every node near coord/.append style={xshift=2.6pt}] coordinates {(1,.344) (2,.076) (3,.105)};
\end{axis}
\end{tikzpicture}
\end{minipage}\hfill
\begin{minipage}[t]{0.47\linewidth}
\centering
\textbf{(b)} External mathematics data\par\smallskip
\begin{tabular*}{\linewidth}{@{\extracolsep{\fill}}lcccc@{}}
\toprule
 & \multicolumn{2}{c}{OpenThoughts-Math} & \multicolumn{2}{c}{DeepMath-103K}\\
\cmidrule(lr){2-3}\cmidrule(lr){4-5}
Target & AUROC & TPR & AUROC & TPR\\
\midrule
Qwen3-1.7B & .916 & .418 & .772 & .164\\
Qwen3-4B & .922 & .576 & .767 & .193\\
Qwen3.5-2B & .913 & .477 & .740 & .135\\
Qwen3.5-4B & .925 & .582 & .884 & .361\\
Gemma-4-E2B & .832 & .207 & .723 & .123\\
\bottomrule
\end{tabular*}
\end{minipage}
\end{table}
% END INLINE TABLE: main_imitation
When auxiliary data from the target's training distribution are limited, we consider synthetic records and an external dataset for reference-model training. For synthesis, Qwen3.5-27B~\citep{qwen2026qwen35} generates new problems and reference answers based on candidate records, without using membership labels. These records support effective attacks in mathematics and medical question answering, while the weaker performance in coding may reflect the difficulty of jointly matching problem requirements and program structure (Table~\ref{tab:opd_mia_imitation}(a)). Replacing OpenThoughts-Math auxiliary data with DeepMath-103K~\citep{he2025deepmath} still yields effective attacks across all five mathematics targets (Table~\ref{tab:opd_mia_imitation}(b)). This shows that reference models can capture membership signals using auxiliary data from another dataset in the same domain.

\section{Related Work}\label{sec:related-work}

\noindent\textbf{Privacy in distillation.}
Distillation can expose membership information about teacher and student training data~\citep{jagielski2023students,zhang2025distillationprivacy}. We study reference-conditioned OPD, where reference answers guide teacher feedback rather than serve as fixed student targets. Our empirical auditing complements private training methods such as DP-OPD~\citep{khadem2026dpopd}.

\noindent\textbf{Membership inference in language models.} Existing attacks exploit token selection~\citep{shi2024mink,zhang2025minkpp} and reference-model calibration~\citep{carlini2022lira,zarifzadeh2024rmia}. Related methods examine token-level evidence through InfoRMIA~\citep{tao2025informia}, local windows through WBC~\citep{chen2026wbc} and directional scoring in EZ-MIA~\citep{ilic2026ezmia}. Other work audits RLVR through sampled generations~\citep{liu2025diba} and studies membership inference without target probabilities~\citep{kaneko2025samia, he2025petal}. In contrast to these approaches, \method{} extracts membership signals from fresh student trajectories through token-wise maximum-reference calibration and asymmetric aggregation.

\section{Conclusion}\label{sec:conclusion}
We study membership leakage in reference-conditioned on-policy distillation and show that fresh student trajectories reveal signals that fixed reference-answer losses can miss. We introduce \method{}, which combines token-level reference-model calibration with asymmetric aggregation and outperforms all evaluated baselines across fifteen targets in three domains. Leakage remains detectable through text-only trajectory-prefix queries, while DP-LoRA reduces \method{}’s AUROC to near chance with smaller utility gains in the evaluated settings. These findings show that excluding reference answers from the student’s training input does not by itself ensure membership privacy, motivating privacy audits that examine generation behavior rather than relying on fixed-answer likelihoods.

\bibliography{references}
\bibliographystyle{references}
% Appendix of main.tex, included there with \input{appendix} after the bibliography.
\clearpage\appendix
\raggedbottom
\numberwithin{figure}{section}
\section{Understanding \method{}}\label{app:residue-method}

\subsection{Role of Reference-Model Calibration}\label{app:residue-shadow-calibration}
OPD changes token predictions through both the candidate record and other training records. The target's probability increase relative to its initialization therefore includes changes shared across records. Reference models trained with the same procedure but without the candidate help account for these shared changes. On the mathematics Qwen3-1.7B target analyzed in the main text, replacing the base-model log-probability with the mean across trained reference models raises AUROC from 0.513 to 0.853, with the candidates, trajectories, and averaging procedure fixed. This comparison supports using trained models for calibration.

\subsection{Negative Gaps as a Correction}\label{app:leaky-negative-correction}
A model trained without the candidate can still exceed all reference models at some positions. The positive contribution may therefore increase for two reasons: membership can give the target an additional lead, and larger disagreement among models can produce a larger chance lead. Keeping only positive gaps includes both contributions in the score. Negative gaps measure how far the target trails the same reference and can help account for the latter contribution.

To describe this effect, consider a fixed prefix and token. As an idealized model, suppose that the log-probabilities of models trained without the candidate satisfy
\begin{equation}
\ell_j=\mu+\sigma Z_j,\qquad \sigma>0,
\end{equation}
where the $Z_j$ are independent and identically distributed. The location $\mu$ and scale $\sigma$ may vary across records and positions, while the standardized distribution remains the same. Let $m_k=\mathbb E[\max_{1\leq j\leq k} Z_j]$, assuming finite first moments. Write $M_K$ for the maximum log-probability among the $K$ reference models and $d=\ell_T-M_K$. The positive gap is the increase in the maximum when the target is added, while the negative magnitude is the target's shortfall from this enlarged maximum:
\begin{align}
\pos{d}&=\max(\ell_T,M_K)-M_K,\\
\pos{-d}&=\max(\ell_T,M_K)-\ell_T.
\end{align}
Their expectations are
\begin{equation}
\mathbb E[\pos{d}]=\sigma(m_{K+1}-m_K),\qquad
\mathbb E[\pos{-d}]=\sigma(m_{K+1}-m_1).
\end{equation}
Both share the scale factor $\sigma$. For a nonzero expected negative magnitude, setting
\begin{equation}
c_K=\frac{m_{K+1}-m_K}{m_{K+1}-m_1}
\quad\text{gives}\quad
\mathbb E[\pos{d}-c_K\pos{-d}]=0.
\label{eq:leaky-chance-ratio}
\end{equation}
Under this model, subtracting a scaled negative contribution removes the mean chance lead associated with model disagreement. This gives the leaky ReLU form and explains a possible correction role for negative gaps. The coefficient depends on both the number of references and the standardized distribution.

The derivation concerns fixed scoring positions. We examine its relevance on the actual target-generated trajectories by letting each reference model in turn act as the target against the other seven. None of these reference models was trained on any candidate. In the medical Qwen3-1.7B experiment, we measure disagreement by averaging the tokenwise standard deviation of reference log-probabilities. The Spearman correlation between the positive score and this disagreement measure, averaged over the eight choices of acting target, decreases from 0.331 to 0.011 after subtracting the negative contribution with weight 0.04. Computing the score and disagreement on separate sets of trajectories gives a similar reduction. These observations support a correction that weakens the tendency of positive scores to increase with model disagreement.

\subsection{Choice of the Negative Slope and Empirical Results}\label{app:opd_mia_alpha_plateau}
Approximating the standardized predictions $Z_j$ by independent standard normal variables gives $c_8\approx0.041$ in Equation~\eqref{eq:leaky-chance-ratio}. This approximation provides a basis for a negative slope of about 0.04. We use $\alpha=0.04$ for all targets.

\begin{figure}[!htbp]\centering% BEGIN INLINE FIGURE: alpha_targets
\definecolor{abBlue}{HTML}{74A9C5}\definecolor{abRose}{HTML}{DD7389}\definecolor{abSand}{HTML}{C4AC5E}%
\pgfplotsset{sm panel/.style={scale only axis,axis lines*=left,axis line style={black!40,line width=.4pt},tick style={black!40,line width=.4pt},tick align=outside,major tick length=2pt,tick label style={font=\scriptsize,text=black!70},label style={font=\scriptsize,text=black!85},xlabel style={text depth=.3ex},ylabel style={align=center},grid style={black!10,line width=.3pt},scaled ticks=false,clip=false,title style={font=\small,yshift=-2pt,text depth=0pt},legend style={font=\scriptsize,draw=none,fill=none,inner sep=1pt,cells={anchor=west},/tikz/every even column/.append style={column sep=7pt}},legend image code/.code={\draw[mark repeat=2,mark phase=2,##1] plot coordinates {(0cm,0cm) (.2cm,0cm) (.4cm,0cm)};}},sm series/.style={color=#1,line width=.9pt,mark=*,mark size=1.6pt,mark options={fill=#1,draw=white,line width=.4pt}},dom math/.style={sm series=abBlue},dom med/.style={sm series=abRose,mark=square*,mark size=1.5pt},dom code/.style={sm series=abSand,mark=triangle*,mark size=2pt},sm dot/.style={only marks,mark options={fill=#1,draw=white,line width=.4pt}},dot math/.style={sm dot=abBlue,mark=*,mark size=1.8pt},dot med/.style={sm dot=abRose,mark=square*,mark size=1.6pt},dot code/.style={sm dot=abSand,mark=triangle*,mark size=2.2pt},fixed2/.style={/pgf/number format/fixed,/pgf/number format/precision=2},fixed1/.style={/pgf/number format/fixed,/pgf/number format/precision=1}}%
\begin{tikzpicture}
\begin{axis}[sm panel,name=r0c0,at={(0,0)},anchor=north west,width=2.2cm,height=2.2cm,xmin=-.008,xmax=.108,xtick={0,.04,.08},x tick label style={fixed2},ymin=.76,ymax=.98,ytick={.8,.85,.9,.95},ymajorgrids,y tick label style={fixed2},title={Qwen3-1.7B},xticklabels={},ylabel={Target-generated\\AUROC}]
\draw[black!35,dashed,line width=.4pt] (axis cs:.04,.76) -- (axis cs:.04,.98);
\addplot[dom math] coordinates {(0,.876) (.01,.8921) (.02,.9052) (.03,.9131) (.04,.9158) (.05,.9148) (.06,.9115) (.08,.9002) (.1,.8865)};
\addplot[dom med] coordinates {(0,.9605) (.01,.9625) (.02,.9645) (.03,.9657) (.04,.9665) (.05,.9667) (.06,.9669) (.08,.9661) (.1,.9645)};
\addplot[dom code] coordinates {(0,.8175) (.01,.8252) (.02,.833) (.03,.8405) (.04,.8467) (.05,.8517) (.06,.8554) (.08,.8583) (.1,.8574)};
\end{axis}
\begin{axis}[sm panel,name=r0c1,at={(r0c0.south east)},anchor=south west,xshift=0.3cm,width=2.2cm,height=2.2cm,xmin=-.008,xmax=.108,xtick={0,.04,.08},x tick label style={fixed2},ymin=.76,ymax=.98,ytick={.8,.85,.9,.95},ymajorgrids,y tick label style={fixed2},title={Qwen3-4B},xticklabels={},yticklabels={}]
\draw[black!35,dashed,line width=.4pt] (axis cs:.04,.76) -- (axis cs:.04,.98);
\addplot[dom math] coordinates {(0,.8809) (.01,.8941) (.02,.9058) (.03,.9154) (.04,.9224) (.05,.927) (.06,.9295) (.08,.9281) (.1,.9233)};
\addplot[dom med] coordinates {(0,.8241) (.01,.8321) (.02,.8387) (.03,.8444) (.04,.8482) (.05,.8503) (.06,.8506) (.08,.8485) (.1,.8439)};
\addplot[dom code] coordinates {(0,.796) (.01,.8125) (.02,.8267) (.03,.8377) (.04,.8434) (.05,.845) (.06,.8435) (.08,.8347) (.1,.822)};
\end{axis}
\begin{axis}[sm panel,name=r0c2,at={(r0c1.south east)},anchor=south west,xshift=0.3cm,width=2.2cm,height=2.2cm,xmin=-.008,xmax=.108,xtick={0,.04,.08},x tick label style={fixed2},ymin=.76,ymax=.98,ytick={.8,.85,.9,.95},ymajorgrids,y tick label style={fixed2},title={Qwen3.5-2B},xticklabels={},yticklabels={}]
\draw[black!35,dashed,line width=.4pt] (axis cs:.04,.76) -- (axis cs:.04,.98);
\addplot[dom math] coordinates {(0,.8986) (.01,.9046) (.02,.9083) (.03,.9114) (.04,.9129) (.05,.9128) (.06,.9119) (.08,.9061) (.1,.8966)};
\addplot[dom med] coordinates {(0,.8586) (.01,.862) (.02,.8645) (.03,.8664) (.04,.8676) (.05,.8682) (.06,.8682) (.08,.8669) (.1,.8643)};
\addplot[dom code] coordinates {(0,.8437) (.01,.8637) (.02,.8828) (.03,.8964) (.04,.9013) (.05,.8976) (.06,.8893) (.08,.8681) (.1,.8458)};
\end{axis}
\begin{axis}[sm panel,name=r0c3,at={(r0c2.south east)},anchor=south west,xshift=0.3cm,width=2.2cm,height=2.2cm,xmin=-.008,xmax=.108,xtick={0,.04,.08},x tick label style={fixed2},ymin=.76,ymax=.98,ytick={.8,.85,.9,.95},ymajorgrids,y tick label style={fixed2},title={Qwen3.5-4B},xticklabels={},yticklabels={}]
\draw[black!35,dashed,line width=.4pt] (axis cs:.04,.76) -- (axis cs:.04,.98);
\addplot[dom math] coordinates {(0,.9122) (.01,.9176) (.02,.9218) (.03,.9245) (.04,.9251) (.05,.9249) (.06,.9237) (.08,.9191) (.1,.913)};
\addplot[dom med] coordinates {(0,.8225) (.01,.8273) (.02,.8316) (.03,.8347) (.04,.837) (.05,.8382) (.06,.839) (.08,.8388) (.1,.8362)};
\addplot[dom code] coordinates {(0,.8015) (.01,.8126) (.02,.8244) (.03,.8361) (.04,.846) (.05,.854) (.06,.8587) (.08,.8612) (.1,.8576)};
\end{axis}
\begin{axis}[sm panel,name=r0c4,at={(r0c3.south east)},anchor=south west,xshift=0.3cm,width=2.2cm,height=2.2cm,xmin=-.008,xmax=.108,xtick={0,.04,.08},x tick label style={fixed2},ymin=.76,ymax=.98,ytick={.8,.85,.9,.95},ymajorgrids,y tick label style={fixed2},title={Gemma-4-E2B},xticklabels={},yticklabels={}]
\draw[black!35,dashed,line width=.4pt] (axis cs:.04,.76) -- (axis cs:.04,.98);
\addplot[dom math] coordinates {(0,.7726) (.01,.7911) (.02,.8082) (.03,.8226) (.04,.832) (.05,.8376) (.06,.8384) (.08,.8306) (.1,.8152)};
\addplot[dom med] coordinates {(0,.8043) (.01,.8115) (.02,.8176) (.03,.8224) (.04,.8257) (.05,.8275) (.06,.8278) (.08,.826) (.1,.8213)};
\addplot[dom code] coordinates {(0,.7827) (.01,.801) (.02,.8181) (.03,.8318) (.04,.8393) (.05,.8414) (.06,.838) (.08,.8253) (.1,.8081)};
\end{axis}
\begin{axis}[sm panel,name=r1c0,at={(r0c0.south west)},anchor=north west,yshift=-0.3cm,width=2.2cm,height=2.2cm,xmin=-.008,xmax=.108,xtick={0,.04,.08},x tick label style={fixed2},ymin=.76,ymax=.98,ytick={.8,.85,.9,.95},ymajorgrids,y tick label style={fixed2},xlabel={Slope $\alpha$},ylabel={Teacher-generated\\AUROC}]
\draw[black!35,dashed,line width=.4pt] (axis cs:.04,.76) -- (axis cs:.04,.98);
\addplot[dom math] coordinates {(0,.8741) (.01,.891) (.02,.9037) (.03,.9111) (.04,.913) (.05,.9108) (.06,.9062) (.08,.8916) (.1,.8753)};
\addplot[dom med] coordinates {(0,.9635) (.01,.9671) (.02,.9698) (.03,.9715) (.04,.9723) (.05,.9727) (.06,.9725) (.08,.9714) (.1,.9694)};
\addplot[dom code] coordinates {(0,.8899) (.01,.897) (.02,.903) (.03,.9077) (.04,.911) (.05,.9137) (.06,.9157) (.08,.9176) (.1,.917)};
\end{axis}
\begin{axis}[sm panel,name=r1c1,at={(r1c0.south east)},anchor=south west,xshift=0.3cm,width=2.2cm,height=2.2cm,xmin=-.008,xmax=.108,xtick={0,.04,.08},x tick label style={fixed2},ymin=.76,ymax=.98,ytick={.8,.85,.9,.95},ymajorgrids,y tick label style={fixed2},xlabel={Slope $\alpha$},yticklabels={}]
\draw[black!35,dashed,line width=.4pt] (axis cs:.04,.76) -- (axis cs:.04,.98);
\addplot[dom math] coordinates {(0,.8885) (.01,.905) (.02,.9193) (.03,.9304) (.04,.9375) (.05,.9412) (.06,.9424) (.08,.9394) (.1,.9324)};
\addplot[dom med] coordinates {(0,.8518) (.01,.862) (.02,.8655) (.03,.8644) (.04,.8592) (.05,.8533) (.06,.8469) (.08,.8344) (.1,.8223)};
\addplot[dom code] coordinates {(0,.8573) (.01,.8663) (.02,.8718) (.03,.8732) (.04,.8725) (.05,.8691) (.06,.8645) (.08,.8523) (.1,.8394)};
\end{axis}
\begin{axis}[sm panel,name=r1c2,at={(r1c1.south east)},anchor=south west,xshift=0.3cm,width=2.2cm,height=2.2cm,xmin=-.008,xmax=.108,xtick={0,.04,.08},x tick label style={fixed2},ymin=.76,ymax=.98,ytick={.8,.85,.9,.95},ymajorgrids,y tick label style={fixed2},xlabel={Slope $\alpha$},yticklabels={},legend style={at={(0.500,-0.520)},anchor=north,legend columns=-1}]
\draw[black!35,dashed,line width=.4pt] (axis cs:.04,.76) -- (axis cs:.04,.98);
\addplot[dom math] coordinates {(0,.9412) (.01,.9484) (.02,.9531) (.03,.9563) (.04,.9576) (.05,.9579) (.06,.9567) (.08,.9519) (.1,.9451)};\addlegendentry{Mathematics}
\addplot[dom med] coordinates {(0,.8827) (.01,.8856) (.02,.8882) (.03,.89) (.04,.8913) (.05,.8917) (.06,.8913) (.08,.8881) (.1,.8838)};\addlegendentry{Medical}
\addplot[dom code] coordinates {(0,.8956) (.01,.908) (.02,.9153) (.03,.9178) (.04,.9162) (.05,.9125) (.06,.9074) (.08,.8932) (.1,.8789)};\addlegendentry{Coding}
\end{axis}
\begin{axis}[sm panel,name=r1c3,at={(r1c2.south east)},anchor=south west,xshift=0.3cm,width=2.2cm,height=2.2cm,xmin=-.008,xmax=.108,xtick={0,.04,.08},x tick label style={fixed2},ymin=.76,ymax=.98,ytick={.8,.85,.9,.95},ymajorgrids,y tick label style={fixed2},xlabel={Slope $\alpha$},yticklabels={}]
\draw[black!35,dashed,line width=.4pt] (axis cs:.04,.76) -- (axis cs:.04,.98);
\addplot[dom math] coordinates {(0,.9444) (.01,.9461) (.02,.9451) (.03,.9419) (.04,.9384) (.05,.934) (.06,.9299) (.08,.9204) (.1,.9097)};
\addplot[dom med] coordinates {(0,.8595) (.01,.8653) (.02,.8701) (.03,.8745) (.04,.878) (.05,.88) (.06,.8809) (.08,.8797) (.1,.8764)};
\addplot[dom code] coordinates {(0,.8285) (.01,.843) (.02,.8571) (.03,.8685) (.04,.8762) (.05,.8804) (.06,.881) (.08,.8759) (.1,.868)};
\end{axis}
\begin{axis}[sm panel,name=r1c4,at={(r1c3.south east)},anchor=south west,xshift=0.3cm,width=2.2cm,height=2.2cm,xmin=-.008,xmax=.108,xtick={0,.04,.08},x tick label style={fixed2},ymin=.76,ymax=.98,ytick={.8,.85,.9,.95},ymajorgrids,y tick label style={fixed2},xlabel={Slope $\alpha$},yticklabels={}]
\draw[black!35,dashed,line width=.4pt] (axis cs:.04,.76) -- (axis cs:.04,.98);
\addplot[dom math] coordinates {(0,.8195) (.01,.8391) (.02,.8532) (.03,.8615) (.04,.8643) (.05,.8634) (.06,.859) (.08,.8448) (.1,.8288)};
\addplot[dom med] coordinates {(0,.8593) (.01,.8676) (.02,.8738) (.03,.8778) (.04,.8805) (.05,.8813) (.06,.8815) (.08,.879) (.1,.8748)};
\addplot[dom code] coordinates {(0,.7933) (.01,.809) (.02,.8214) (.03,.8288) (.04,.8305) (.05,.8275) (.06,.821) (.08,.8044) (.1,.7849)};
\end{axis}
\end{tikzpicture}
% END INLINE FIGURE: alpha_targets
\caption{Gray-box AUROC against the rectifier slope $\alpha$ for each target, on target-generated (top) and teacher-generated (bottom) trajectories; the dashed line marks $\alpha=0.04$.}\label{fig:opd_alpha_targets}\end{figure}
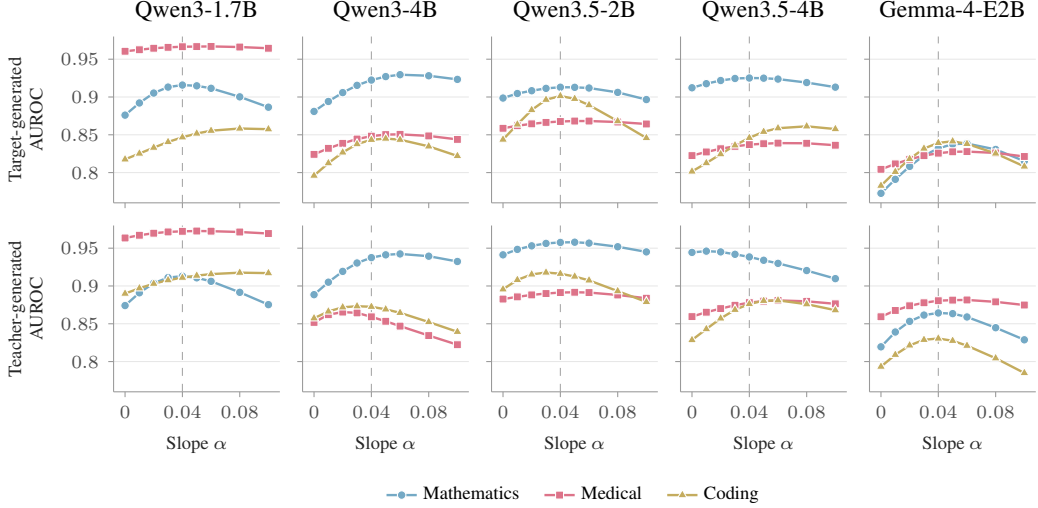

We evaluate the effect of the slope while keeping the candidates, sampled trajectories, reference models, and token scores fixed. Figure~\ref{fig:opd_alpha_targets} shows the results for each model and domain. On target-generated trajectories, introducing the negative contribution raises mean AUROC over the fifteen targets from 0.843 at $\alpha=0$ to 0.875 at $\alpha=0.04$, close to the highest tested mean of 0.877 at $\alpha=0.05$.

On target-generated trajectories, every tested slope in $\{0.04,0.05,0.06,0.08\}$ outperforms $\alpha=0$ on all fifteen targets. The best tested slope varies across targets, but the improvement extends over neighboring values. These results support using a common small negative slope and show that the gain does not depend on choosing exactly 0.04. The figure also reports teacher-generated trajectories as a supplementary comparison.

\section{Experimental Details}\label{app:opd_mia_protocols}
\subsection{Datasets and Membership Evaluation}\label{app:opd_mia_cohorts}
We evaluate membership inference on mathematics, medical question answering, and code generation. Each candidate record $z=(x,r)$ contains a task input $x$ and a reference answer $r$. Membership is determined by whether the record was included in the target's OPD training set. For each evaluation, we initially construct a candidate pool of 1,024 records, containing 512 members and 512 non-members. All evaluated candidates are excluded from reference-model training. Comparisons between attack methods on the same target use the same candidate set.

\noindent\textbf{Mathematics.} We use OpenThoughts-Math, with 29,434 mathematical problems and their reference solutions. The problem statement forms $x$, and the corresponding solution forms $r$. Each target is trained on 512 problems, all of which are included as members in the attack evaluation. We sample 512 non-members from the remaining problems, excluding records used for target or reference-model training. This gives a balanced initial candidate pool of 1,024 problem--solution pairs.

\noindent\textbf{Medical question answering.} We use the training split of MedMCQA. We retain questions with a single correct answer among four options and an explanation of at least 20 words, and remove duplicate questions. The input $x$ contains the question, the four options labeled A--D, and an instruction to return the correct option letter. The reference answer $r$ contains the provided explanation followed by the correct option. Each target is trained on 16,384 records, including the 512 evaluated members. We select 512 non-members from the same filtered dataset, excluding records used for target or reference-model training. All medical targets use the same member and non-member candidates.

\noindent\textbf{Code generation.} We use TACO, which contains programming problems collected from multiple sources, including Codeforces and AtCoder. The input $x$ is the problem statement, and the reference answer $r$ is the associated Python solution formatted as a code block. Each target is trained on 8,339 records. We select 512 members from the training set and 512 non-members from outside it. To account for differences in problem composition, we divide records into strata according to their source, whether a function name is specified, and problem-statement length. Length is measured in characters and divided into three groups using the training-pool terciles. We select equal numbers of members and non-members within each stratum and match their available difficulty labels. This construction balances the two groups with respect to source, interface format, length group, and annotated difficulty.

\subsection{Baselines}\label{app:opd_mia_baselines}

\noindent We evaluate all baselines on the same candidates as \method. \textbf{LOSS}~\citep{yeom2018privacy} uses the mean token-level loss on the reference answer, while \textbf{ZLIB}~\citep{carlini2021extracting} normalizes the total loss by the compressed byte length of the reference answer. \textbf{Min-K\%}~\citep{shi2024mink} averages the lowest 20\% of token log-probabilities. \textbf{Min-K\%++}~\citep{zhang2025minkpp} uses the same selection ratio after standardizing each token's log-probability using the corresponding vocabulary distribution.

\noindent\textbf{Ratio}~\citep{watson2022difficulty}, \textbf{WBC}~\citep{chen2026wbc}, and \textbf{EZ-MIA}~\citep{ilic2026ezmia} use the frozen checkpoint that initializes the target as the reference model. Ratio measures the reduction in reference-answer loss relative to this model. WBC compares target and reference losses over sliding windows, using ten window sizes spanning 2--40 tokens and averaging the fraction of windows in which the target has lower loss. EZ-MIA restricts the comparison to positions where the target's greedy prediction differs from the reference token and computes the ratio of positive to negative target-minus-reference log-probability differences.

\noindent\textbf{LiRA}~\citep{carlini2022lira}, \textbf{RMIA}~\citep{zarifzadeh2024rmia}, and \textbf{InfoRMIA}~\citep{tao2025informia} use the same eight reference models as \method, trained without the evaluated candidates. We use offline LiRA with the loss statistic, standardized by its mean and standard deviation across reference models. We use offline RMIA with $\gamma=1$. For InfoRMIA, the reference distribution is the arithmetic mean of the reference probability distributions; we use the token-level score with mean aggregation across tokens.

\noindent\textbf{ICP-MIA}~\citep{lu2026icpmia} uses the reference-probe variant with ten probes per candidate. We retrieve problem--answer pairs from a held-out pool using cosine similarity between all-MiniLM-L6-v2 embeddings, excluding the evaluated candidates. Each probe is added separately as an in-context demonstration, and the attack measures the resulting change in the target's reference-answer log-probability.

\noindent\textbf{DIBA}~\citep{liu2025diba} samples eight trajectories per candidate from each of the target and base models at temperature 0.5. It combines correctness features with features measuring differences in the models' generation behavior, using a random-forest and logistic-regression stacking classifier. We train the classifier with membership labels under five-fold stratified cross-validation and compute attack results from pooled out-of-fold predictions. For coding, the two correctness features are fixed to zero, leaving five varying features.

\subsection{Models and Training}\label{app:opd_mia_training}

\noindent\textbf{Student--teacher pairings.} We evaluate five students from three model families: Qwen3-1.7B and Qwen3-4B~\citep{qwen2025qwen3}, Qwen3.5-2B and Qwen3.5-4B~\citep{qwen2026qwen35}, and Gemma-4-E2B~\citep{gemma2026gemma4}. Each student is trained separately on mathematics, medical question answering, and code generation, giving fifteen target models. Each target uses a larger teacher from the same model family. Table~\ref{tab:opd_pairings} lists the student--teacher pairings for each domain.

% BEGIN INLINE TABLE: training_pairings
\begin{table}[!htbp]
\centering\small\setlength{\tabcolsep}{3pt}\renewcommand{\arraystretch}{1.15}
\caption{Student and teacher models used for OPD in each domain.}\label{tab:opd_pairings}
\begin{tabular*}{\linewidth}{@{\extracolsep{\fill}}l*{3}{l}@{}}
\toprule
Student & Mathematics teacher & Medical teacher & Coding teacher\\\midrule
Qwen3-1.7B & Qwen3-8B & Qwen3-8B & Qwen3-8B\\
Qwen3-4B & Qwen3-14B & Qwen3-14B & Qwen3-14B\\
Qwen3.5-2B & Qwen3.5-27B & Qwen3.5-9B & Qwen3.5-9B\\
Qwen3.5-4B & Qwen3.5-27B & Qwen3.5-27B & Qwen3.5-27B\\
Gemma-4-E2B & Gemma-4-26B-A4B & Gemma-4-26B-A4B & Gemma-4-26B-A4B\\
\bottomrule
\end{tabular*}
\end{table}
% END INLINE TABLE: training_pairings

\noindent\textbf{OPD training.} All targets in Table~\ref{tab:opd_main} use full-parameter fine-tuning with a frozen teacher. The student receives the problem $x$, while the teacher receives both $x$ and the reference answer $r$. During training, the student samples trajectories from its current policy at temperature 1.1 and learns from the teacher's next-token distributions on these trajectories. The distillation loss uses a generalized Jensen--Shannon divergence with interpolation weight $\beta$ and per-token clipping. Mathematics experiments use 126 training steps with a global batch size of 32. Medical and coding experiments use 512 training steps with a global batch size of 128. Table~\ref{tab:opd_recipe} summarizes the training configurations. All teacher models use thinking mode, while student models use non-thinking mode.

\opdLong{OPD training configurations. Training steps indicate the total training budget.}{tab:opd_recipe}{@{\extracolsep{\fill}}l*{4}{c}@{}}
{Student & \makecell{Interpolation\\weight ($\beta$)} & \makecell{Learning\\rate} & \makecell{Training\\steps} & \makecell{Global\\batch size}\\\midrule}
{\multicolumn{5}{@{}l}{\textit{Mathematics}}\\
Qwen3-1.7B & 0 & 5e-6 & 126 & 32\\
Qwen3-4B & 0 & 2e-6 & 126 & 32\\
Qwen3.5-2B & 0.5 & 1e-6 & 126 & 32\\
Qwen3.5-4B & 0.5 & 1e-6 & 126 & 32\\
Gemma-4-E2B & 0 & 3e-6 & 126 & 32\\
\multicolumn{5}{@{}l}{\textit{Medical question answering}}\\
Qwen3-1.7B & 0 & 1e-5 & 512 & 128\\
Qwen3-4B & 0 & 4e-6 & 512 & 128\\
Qwen3.5-2B & 0.5 & 2e-6 & 512 & 128\\
Qwen3.5-4B & 0.5 & 1e-6 & 512 & 128\\
Gemma-4-E2B & 0.5 & 5e-6 & 512 & 128\\
\multicolumn{5}{@{}l}{\textit{Code generation}}\\
Qwen3-1.7B & 0 & 2e-6 & 512 & 128\\
Qwen3-4B & 0 & 2e-6 & 512 & 128\\
Qwen3.5-2B & 0.5 & 1e-6 & 512 & 128\\
Qwen3.5-4B & 0.5 & 2e-6 & 512 & 128\\
Gemma-4-E2B & 0.5 & 5e-6 & 512 & 128\\}{5}
% END INLINE TABLE: training_recipe

\subsection{\method{} Implementation}\label{app:opd_mia_primary_protocol}

\noindent\textbf{Trajectory sampling.} For each candidate record $z=(x,r)$, we sample eight trajectories from the target model conditioned on the problem $x$ and reference answer $r$. We use a sampling temperature of 1.1, top-$p$ of 0.95, and top-$k$ of 20. These trajectories are generated at attack time.

\noindent\textbf{Token scoring and aggregation.} We score each sampled token using the target and eight reference models, conditioned on the same problem $x$ and trajectory prefix. The reference answer $r$ is used for sampling and is not separately appended to the scoring input. Token log-probabilities are computed at temperature 1.1. We compute the membership score using the leaky ReLU aggregation defined in Section~\ref{sec:residue-attack}, first averaging over all scored tokens within each trajectory and then averaging the trajectory scores with equal weight. We use the same negative slope $\alpha=0.04$ for all targets. Appendix~\ref{app:opd_mia_alpha_plateau} explains this choice and reports the sensitivity to $\alpha$.

\noindent\textbf{Blackbox implementation.}
The black-box setting uses one greedy next-token query per scored prefix, giving $Q(z)=\sum_i L_i$ scoring queries per candidate. We score up to 1,024 positions per trajectory for Qwen3 and all coding targets, and up to 512 for Qwen3.5 and Gemma-4 mathematics and medical targets. With eight trajectories, the mean number of scoring queries per candidate ranges from 4,045 to 8,192 across targets; these counts exclude trajectory generation and local reference-model computation. Appendix~\ref{app:opd_mia_blackbox} defines the score and reports the results.

\subsection{Downstream Utility}\label{app:opd_mia_utility}

\opdLong{Downstream utility (Avg@8, \%) of the evaluated targets; $\Delta$ is the change from the base in percentage points. Training steps indicate the total training budget.}{tab:opd_utility}{@{\extracolsep{\fill}}l*{4}{c}@{}}
{Student & \makecell{Training\\steps} & Base & Trained & $\Delta$ (pp)\\\midrule}
{\multicolumn{5}{@{}l}{\textit{Mathematics: AIME 2024}}\\
Qwen3-1.7B & 126 & 47.50 & 52.50 & +5.00\\
Qwen3-4B & 126 & 72.92 & 75.00 & +2.08\\
Qwen3.5-2B & 126 & 31.25 & 42.50 & +11.25\\
Qwen3.5-4B & 126 & 85.00 & 87.08 & +2.08\\
Gemma-4-E2B & 126 & 41.93 & 45.00 & +3.07\\
\multicolumn{5}{@{}l}{\textit{Medical QA: MedQA-USMLE}}\\
Qwen3-1.7B & 512 & 49.00 & 52.98 & +3.98\\
Qwen3-4B & 512 & 65.63 & 70.38 & +4.75\\
Qwen3.5-2B & 512 & 60.17 & 66.01 & +5.84\\
Qwen3.5-4B & 512 & 83.77 & 86.69 & +2.92\\
Gemma-4-E2B & 512 & 63.58 & 65.00 & +1.42\\
\multicolumn{5}{@{}l}{\textit{Code generation: LiveCodeBench v6}}\\
Qwen3-1.7B & 512 & 31.00 & 31.57 & +0.57\\
Qwen3-4B & 512 & 45.79 & 46.71 & +0.92\\
Qwen3.5-2B & 512 & 20.14 & 22.79 & +2.65\\
Qwen3.5-4B & 512 & 55.43 & 57.57 & +2.14\\
Gemma-4-E2B & 512 & 43.50 & 44.57 & +1.07\\}{5}

\opdLong{Utility decoding per base/target pair (not used for attack scoring). Coding uses SGLang~\citep{zheng2024sglang} for Qwen3-1.7B and Qwen3.5 and vLLM~\citep{kwon2023vllm} otherwise.}{tab:opd_utility_decoding}{@{\extracolsep{\fill}}llr@{}}
{Model family & Decoding & Output tokens\\\midrule}
{\multicolumn{3}{@{}l}{\textit{Mathematics: reasoning enabled; 30 AIME questions}}\\
Qwen3 & $T=1.0$, top-$p=0.95$, no top-$k$ & 38,912\\
Qwen3.5 & $T=1.0$, top-$p=0.95$, top-$k=20$, presence penalty 1.5 & 81,920\\
Gemma-4 & $T=1.0$, top-$p=0.95$, top-$k=64$ & 32,768\\
\multicolumn{3}{@{}l}{\textit{Medical: reasoning disabled; 1,273 MedQA-USMLE questions}}\\
All models & $T=0.7$, top-$p=0.8$, top-$k=20$, presence penalty 1.5 & 32,768\\
\multicolumn{3}{@{}l}{\textit{Coding: reasoning enabled; 175 LiveCodeBench v6 questions}}\\
Qwen3 & $T=0.6$, top-$p=0.95$, top-$k=20$ & 32,768\\
Qwen3.5 & $T=1.0$, top-$p=0.95$, top-$k=20$, presence penalty 1.5 & 81,920\\
Gemma-4 & $T=1.0$, top-$p=0.95$, top-$k=64$ & 32,768\\}{3}
% END INLINE TABLE: utility_decoding

\subsection{Hardware and Software}\label{app:opd_mia_compute}
\noindent\textbf{Hardware.} All experiments run on single servers of an NVIDIA GB200 NVL72 system. Each server has four NVIDIA B200 GPUs with 184\,GiB of HBM3e memory each, connected by NVLink, two 72-core NVIDIA Grace CPUs, and about 1.5\,TB of host memory. OPD training of targets and reference models, including the LoRA and DP-LoRA jobs, uses all four GPUs of a server, with the teacher served on the same GPUs. Trajectory sampling, scoring, and downstream utility evaluation use one to four GPUs per job.

\noindent\textbf{Software.} Jobs run in the NVIDIA NGC vLLM containers (releases 25.12 and 26.07; Python 3.12). OPD training builds on the ReNIO implementation~\citep{lin2026renio} with Transformers 4.57.1, TRL 0.26.0, Accelerate 1.11.0 in bf16 mixed precision, and DeepSpeed 0.18.2; the teacher's next-token distributions are computed with vLLM~\citep{kwon2023vllm}. LoRA uses PEFT 0.17.1, and DP-LoRA uses Opacus 1.6.0~\citep{yousefpour2021opacus} for per-record clipping, noise addition, and privacy accounting. Trajectory sampling and token log-probability scoring for \method{} and all baselines use vLLM. Downstream utility decoding uses vLLM or SGLang 0.5.18~\citep{zheng2024sglang} as listed in Table~\ref{tab:opd_utility_decoding}.

\Needspace{8\baselineskip}
\section{Additional Results and Ablations}\label{app:opd_mia_results}
\subsection{Additional Attack Results}
We further evaluate attack performance at 5\% and 10\% FPR to complement the results at 1\% FPR in the main table. As shown in Table~\ref{tab:opd_baselines_all}, \method{} achieves the highest TPR among the evaluated methods on all fifteen targets at both FPR levels. Its advantage thus remains under more relaxed false-positive constraints.

\opdLong
{TPR at 5\% and 10\% FPR for the methods in Table~\ref{tab:opd_main}. Leading zeros omitted; Q3/Q3.5: Qwen3/Qwen3.5; G4: Gemma-4-E2B.}
{tab:opd_baselines_all}
{@{\extracolsep{\fill}}>{\fontsize{8}{9.2}\selectfont}l*{15}{@{\hspace{3pt}}>{\fontsize{8}{9.2}\selectfont}c}@{}}
{ & \multicolumn{5}{c}{\fontsize{8}{9.2}\selectfont Mathematics} & \multicolumn{5}{c}{\fontsize{8}{9.2}\selectfont Medical QA} & \multicolumn{5}{c}{\fontsize{8}{9.2}\selectfont Code generation}\\
\cmidrule(lr){2-6}\cmidrule(lr){7-11}\cmidrule(lr){12-16}
Method & \makecell{Q3\\1.7B} & \makecell{Q3\\4B} & \makecell{Q3.5\\2B} & \makecell{Q3.5\\4B} & \makecell{G4\\E2B} & \makecell{Q3\\1.7B} & \makecell{Q3\\4B} & \makecell{Q3.5\\2B} & \makecell{Q3.5\\4B} & \makecell{G4\\E2B} & \makecell{Q3\\1.7B} & \makecell{Q3\\4B} & \makecell{Q3.5\\2B} & \makecell{Q3.5\\4B} & \makecell{G4\\E2B}\\\midrule}
{\multicolumn{16}{c}{\fontsize{8}{9.2}\selectfont\textit{TPR at 5\% FPR}}\\\midrule
LOSS & .045 & .035 & .049 & .070 & .055 & .039 & .053 & .047 & .049 & .041 & .033 & .043 & .033 & .035 & .047\\
ZLIB & .043 & .039 & .059 & .051 & .045 & .051 & .049 & .041 & .051 & .049 & .049 & .057 & .055 & .082 & .082\\
Min-K\% & .043 & .039 & .049 & .070 & .055 & .043 & .035 & .043 & .039 & .029 & .031 & .037 & .041 & .045 & .053\\
Min-K\%++ & .031 & .037 & .062 & .047 & .053 & .047 & .041 & .059 & .053 & .061 & .035 & .037 & .035 & .061 & .057\\
ICP-MIA & .047 & .074 & .029 & .043 & .035 & .047 & .045 & .035 & .064 & .047 & .072 & .027 & .037 & .031 & .029\\
Ratio & .039 & .051 & .043 & .035 & .039 & .041 & .033 & .039 & .047 & .045 & .039 & .021 & .045 & .070 & .057\\
WBC & .049 & .053 & .016 & .031 & .049 & .045 & .041 & .064 & .061 & .033 & .023 & .023 & .086 & .139 & .031\\
EZ-MIA & .029 & .047 & .037 & .047 & .047 & .049 & .064 & .041 & .053 & .041 & .039 & .064 & .102 & .068 & .000\\
DIBA & .039 & .066 & .035 & .049 & .035 & .088 & .104 & .107 & .055 & .072 & .064 & .066 & .057 & .090 & .064\\
\midrule
LiRA & .180 & .168 & .084 & .078 & .053 & .260 & .141 & .160 & .125 & .045 & .072 & .102 & .117 & .117 & .078\\
RMIA & .189 & .191 & .111 & .066 & .047 & .273 & .119 & .125 & .156 & .031 & .064 & .092 & .070 & .115 & .062\\
InfoRMIA & .172 & .170 & .104 & .104 & .068 & .311 & .164 & .123 & .125 & .057 & .070 & .086 & .098 & .117 & .078\\
\midrule
\method{} & .662 & .693 & .701 & .762 & .434 & .865 & .559 & .646 & .641 & .508 & .568 & .572 & .689 & .643 & .467\\
\midrule
\multicolumn{16}{c}{\fontsize{8}{9.2}\selectfont\textit{TPR at 10\% FPR}}\\\midrule
LOSS & .078 & .070 & .123 & .107 & .111 & .107 & .105 & .107 & .113 & .076 & .072 & .090 & .070 & .088 & .094\\
ZLIB & .105 & .088 & .104 & .115 & .104 & .090 & .100 & .088 & .082 & .102 & .119 & .170 & .129 & .160 & .178\\
Min-K\% & .070 & .074 & .117 & .111 & .111 & .094 & .092 & .078 & .104 & .086 & .090 & .096 & .074 & .092 & .084\\
Min-K\%++ & .098 & .096 & .129 & .102 & .092 & .102 & .090 & .119 & .121 & .094 & .082 & .062 & .080 & .094 & .098\\
ICP-MIA & .107 & .158 & .084 & .107 & .094 & .121 & .123 & .104 & .113 & .090 & .109 & .084 & .082 & .082 & .070\\
Ratio & .102 & .104 & .086 & .076 & .066 & .090 & .072 & .086 & .102 & .113 & .062 & .055 & .086 & .104 & .119\\
WBC & .096 & .078 & .057 & .072 & .098 & .096 & .080 & .104 & .105 & .090 & .059 & .045 & .154 & .195 & .062\\
EZ-MIA & .068 & .086 & .082 & .076 & .092 & .102 & .113 & .113 & .117 & .107 & .061 & .117 & .158 & .123 & .000\\
DIBA & .102 & .098 & .078 & .080 & .076 & .211 & .205 & .193 & .119 & .121 & .107 & .107 & .100 & .168 & .117\\
\midrule
LiRA & .287 & .242 & .154 & .137 & .094 & .363 & .223 & .256 & .229 & .088 & .133 & .158 & .166 & .207 & .131\\
RMIA & .295 & .293 & .154 & .133 & .072 & .379 & .213 & .248 & .229 & .104 & .129 & .186 & .168 & .188 & .109\\
InfoRMIA & .268 & .271 & .174 & .160 & .105 & .391 & .250 & .201 & .201 & .125 & .141 & .184 & .156 & .229 & .170\\
\midrule
\method{} & .758 & .799 & .805 & .811 & .557 & .928 & .666 & .703 & .686 & .600 & .633 & .643 & .752 & .697 & .590\\}
{16}

\Needspace{5\baselineskip}
\subsection{Baselines and Scoring Rules on the Sampled Trajectories}\label{app:opd_mia_aggregation}
We apply the baselines and several scoring rules to the same candidates and eight target-generated trajectories used in Table~\ref{tab:opd_main}. Each statistic is computed per trajectory, under the student prompt and scoring temperature of Table~\ref{tab:opd_main}, and then averaged over the eight trajectory scores. This comparison examines whether scoring the sampled trajectories is sufficient, and how reference models and token aggregation affect membership inference.

\noindent\textbf{Baselines on the sampled trajectories.}
Table~\ref{tab:opd_cont_baselines} reports the baselines of Table~\ref{tab:opd_main} on the trajectories instead of the reference answer; DIBA, which generates its own rollouts, is not included. Scores without OPD-trained reference models remain close to chance on the trajectories (AUROC 0.47--0.61), whereas LiRA, RMIA, and InfoRMIA gain substantially from trajectory scoring: InfoRMIA rises from 0.740 to 0.949 on medical Qwen3-1.7B and becomes the strongest baseline, with a mean AUROC of 0.826 over the fifteen targets. \method{} still exceeds InfoRMIA on every target, by 0.004--0.100 AUROC (mean 0.049), and has the highest TPR at 1\%, 5\% and 10\% FPR on all fifteen targets.

\opdLong
{Baselines applied to the eight target-generated trajectories that \method{} scores, on the candidates of Table~\ref{tab:opd_main}: AUROC and TPR at 1\%, 5\% and 10\% FPR, leading zeros omitted, best per column in bold. Q3/Q3.5: Qwen3/Qwen3.5; G4: Gemma-4-E2B.}
{tab:opd_cont_baselines}
{@{\extracolsep{\fill}}>{\fontsize{8}{9.2}\selectfont}l*{15}{@{\hspace{3pt}}>{\fontsize{8}{9.2}\selectfont}c}@{}}
{ & \multicolumn{5}{c}{\fontsize{8}{9.2}\selectfont Mathematics} & \multicolumn{5}{c}{\fontsize{8}{9.2}\selectfont Medical QA} & \multicolumn{5}{c}{\fontsize{8}{9.2}\selectfont Code generation}\\
\cmidrule(lr){2-6}\cmidrule(lr){7-11}\cmidrule(lr){12-16}
Method & \makecell{Q3\\1.7B} & \makecell{Q3\\4B} & \makecell{Q3.5\\2B} & \makecell{Q3.5\\4B} & \makecell{G4\\E2B} & \makecell{Q3\\1.7B} & \makecell{Q3\\4B} & \makecell{Q3.5\\2B} & \makecell{Q3.5\\4B} & \makecell{G4\\E2B} & \makecell{Q3\\1.7B} & \makecell{Q3\\4B} & \makecell{Q3.5\\2B} & \makecell{Q3.5\\4B} & \makecell{G4\\E2B}\\\midrule}
{\multicolumn{16}{c}{\fontsize{8}{9.2}\selectfont\textit{AUROC}}\\\midrule
LOSS & .502 & .477 & .489 & .506 & .495 & .536 & .549 & .568 & .557 & .550 & .534 & .536 & .547 & .551 & .544\\
ZLIB & .504 & .472 & .490 & .507 & .487 & .505 & .530 & .558 & .546 & .540 & .531 & .543 & .552 & .555 & .553\\
Min-K\% & .501 & .473 & .492 & .510 & .496 & .531 & .548 & .572 & .559 & .544 & .530 & .536 & .557 & .556 & .545\\
Min-K\%++ & .486 & .472 & .491 & .492 & .495 & .491 & .482 & .564 & .550 & .499 & .510 & .503 & .558 & .566 & .512\\
ICP-MIA & .531 & .511 & .521 & .526 & .502 & .576 & .525 & .570 & .540 & .547 & .531 & .525 & .591 & .576 & .563\\
Ratio & .513 & .565 & .583 & .559 & .533 & .492 & .491 & .539 & .511 & .491 & .513 & .477 & .506 & .516 & .519\\
WBC & .508 & .545 & .565 & .606 & .520 & .504 & .505 & .597 & .582 & .479 & .520 & .488 & .528 & .581 & .503\\
EZ-MIA & .503 & .529 & .522 & .519 & .503 & .514 & .545 & .583 & .570 & .531 & .503 & .546 & .542 & .566 & .524\\
\midrule
LiRA & .851 & .856 & .814 & .872 & .717 & .935 & .841 & .839 & .790 & .779 & .714 & .774 & .731 & .765 & .752\\
RMIA & .849 & .843 & .729 & .744 & .711 & .933 & .842 & .835 & .790 & .777 & .711 & .753 & .705 & .764 & .749\\
InfoRMIA & .874 & .877 & .835 & .905 & .746 & .949 & .844 & .841 & .803 & .784 & .762 & .801 & .802 & .792 & .777\\
\midrule
\textbf{\method{}} & \textbf{.916} & \textbf{.922} & \textbf{.913} & \textbf{.925} & \textbf{.832} & \textbf{.966} & \textbf{.848} & \textbf{.868} & \textbf{.837} & \textbf{.826} & \textbf{.847} & \textbf{.843} & \textbf{.901} & \textbf{.846} & \textbf{.839}\\
\midrule
\multicolumn{16}{c}{\fontsize{8}{9.2}\selectfont\textit{TPR at 1\% FPR}}\\\midrule
LOSS & .004 & .008 & .004 & .008 & .008 & .023 & .006 & .016 & .012 & .004 & .012 & .014 & .004 & .004 & .012\\
ZLIB & .012 & .006 & .010 & .008 & .006 & .025 & .008 & .021 & .012 & .006 & .020 & .008 & .016 & .006 & .025\\
Min-K\% & .010 & .010 & .006 & .012 & .008 & .008 & .006 & .020 & .012 & .002 & .020 & .004 & .006 & .004 & .014\\
Min-K\%++ & .006 & .002 & .006 & .002 & .006 & .020 & .000 & .025 & .029 & .012 & .021 & .004 & .014 & .012 & .033\\
ICP-MIA & .016 & .025 & .018 & .020 & .006 & .021 & .008 & .012 & .012 & .020 & .006 & .012 & .016 & .012 & .018\\
Ratio & .000 & .023 & .045 & .012 & .018 & .008 & .014 & .023 & .008 & .006 & .004 & .020 & .012 & .004 & .008\\
WBC & .008 & .027 & .039 & .031 & .021 & .018 & .014 & .037 & .020 & .014 & .004 & .014 & .010 & .025 & .014\\
EZ-MIA & .008 & .016 & .018 & .008 & .014 & .012 & .018 & .020 & .014 & .008 & .004 & .016 & .006 & .016 & .023\\
\midrule
LiRA & .352 & .289 & .207 & .258 & .113 & .486 & .352 & .400 & .283 & .193 & .174 & .141 & .172 & .342 & .180\\
RMIA & .373 & .361 & .162 & .215 & .135 & .572 & .373 & .426 & .299 & .207 & .184 & .139 & .162 & .326 & .172\\
InfoRMIA & .334 & .418 & .359 & .516 & .135 & .613 & .383 & .451 & .375 & .270 & .242 & .246 & .299 & .355 & .250\\
\midrule
\textbf{\method{}} & \textbf{.418} & \textbf{.576} & \textbf{.477} & \textbf{.582} & \textbf{.207} & \textbf{.740} & \textbf{.449} & \textbf{.512} & \textbf{.461} & \textbf{.350} & \textbf{.434} & \textbf{.396} & \textbf{.578} & \textbf{.527} & \textbf{.270}\\
\midrule
\multicolumn{16}{c}{\fontsize{8}{9.2}\selectfont\textit{TPR at 5\% FPR}}\\\midrule
LOSS & .051 & .037 & .057 & .035 & .053 & .082 & .055 & .084 & .078 & .068 & .084 & .088 & .062 & .064 & .092\\
ZLIB & .049 & .045 & .049 & .049 & .043 & .072 & .045 & .082 & .076 & .035 & .098 & .094 & .059 & .068 & .072\\
Min-K\% & .051 & .027 & .053 & .041 & .051 & .080 & .051 & .092 & .068 & .055 & .068 & .064 & .068 & .061 & .084\\
Min-K\%++ & .053 & .049 & .027 & .053 & .023 & .053 & .066 & .080 & .062 & .043 & .057 & .043 & .076 & .068 & .061\\
ICP-MIA & .059 & .070 & .061 & .059 & .055 & .090 & .023 & .074 & .051 & .055 & .053 & .066 & .062 & .082 & .078\\
Ratio & .029 & .094 & .094 & .066 & .086 & .061 & .055 & .094 & .047 & .041 & .041 & .061 & .080 & .062 & .053\\
WBC & .055 & .080 & .076 & .076 & .070 & .053 & .047 & .117 & .076 & .053 & .039 & .061 & .041 & .064 & .055\\
EZ-MIA & .072 & .080 & .047 & .059 & .072 & .059 & .047 & .098 & .074 & .053 & .047 & .051 & .047 & .061 & .072\\
\midrule
LiRA & .490 & .508 & .432 & .480 & .225 & .732 & .510 & .535 & .473 & .387 & .297 & .348 & .340 & .434 & .295\\
RMIA & .518 & .502 & .242 & .355 & .230 & .729 & .553 & .535 & .465 & .383 & .307 & .365 & .275 & .443 & .297\\
InfoRMIA & .559 & .586 & .506 & .668 & .312 & .752 & .514 & .549 & .516 & .400 & .369 & .436 & .441 & .498 & .371\\
\midrule
\textbf{\method{}} & \textbf{.662} & \textbf{.693} & \textbf{.701} & \textbf{.762} & \textbf{.434} & \textbf{.865} & \textbf{.559} & \textbf{.646} & \textbf{.641} & \textbf{.508} & \textbf{.568} & \textbf{.572} & \textbf{.689} & \textbf{.643} & \textbf{.467}\\
\midrule
\multicolumn{16}{c}{\fontsize{8}{9.2}\selectfont\textit{TPR at 10\% FPR}}\\\midrule
LOSS & .088 & .076 & .104 & .100 & .102 & .133 & .137 & .166 & .145 & .127 & .146 & .154 & .127 & .123 & .146\\
ZLIB & .105 & .102 & .098 & .115 & .088 & .131 & .121 & .139 & .125 & .121 & .154 & .146 & .125 & .129 & .174\\
Min-K\% & .102 & .070 & .119 & .107 & .102 & .148 & .141 & .156 & .141 & .113 & .145 & .152 & .127 & .139 & .143\\
Min-K\%++ & .117 & .092 & .094 & .094 & .076 & .094 & .098 & .135 & .129 & .074 & .094 & .094 & .113 & .129 & .088\\
ICP-MIA & .117 & .119 & .104 & .129 & .096 & .145 & .107 & .162 & .109 & .137 & .105 & .119 & .143 & .117 & .152\\
Ratio & .092 & .141 & .164 & .174 & .154 & .113 & .098 & .152 & .092 & .107 & .098 & .105 & .107 & .109 & .102\\
WBC & .094 & .143 & .156 & .152 & .125 & .123 & .113 & .207 & .174 & .105 & .104 & .092 & .115 & .143 & .121\\
EZ-MIA & .115 & .127 & .084 & .113 & .125 & .102 & .135 & .174 & .150 & .109 & .102 & .119 & .146 & .104 & .123\\
\midrule
LiRA & .576 & .594 & .557 & .633 & .385 & .826 & .605 & .633 & .545 & .457 & .418 & .459 & .418 & .508 & .432\\
RMIA & .594 & .611 & .363 & .434 & .391 & .822 & .646 & .625 & .547 & .479 & .422 & .451 & .389 & .533 & .441\\
InfoRMIA & .674 & .658 & .602 & .758 & .410 & .857 & .604 & .643 & .609 & .525 & .496 & .529 & .525 & .559 & .496\\
\midrule
\textbf{\method{}} & \textbf{.758} & \textbf{.799} & \textbf{.805} & \textbf{.811} & \textbf{.557} & \textbf{.928} & \textbf{.666} & \textbf{.703} & \textbf{.686} & \textbf{.600} & \textbf{.633} & \textbf{.643} & \textbf{.752} & \textbf{.697} & \textbf{.590}\\}
{16}
% END INLINE TABLE: cont_baselines

\noindent\textbf{Scoring rules.}
Table~\ref{tab:opd_scoring_rules} contains four groups. The first uses the target's log-probabilities: their mean, Min-K\%, and Min-K\%++, with the latter two selecting the lowest 20\% of token scores. The second averages signed gaps using three calibration choices: the frozen base model and the mean or maximum log-probability across the trained reference models. The third and fourth keep the reference mean and the reference maximum, respectively, and change the aggregation, so that the two trained-reference calibrations are crossed with the same aggregations. Min-20\% gap averages the lowest 20\% of signed gaps within each trajectory; unlike Min-K\%, it selects tokens by calibrated gaps rather than target log-probabilities. ReLU retains only positive gaps, while \method{} also includes negative gaps with the small weight used in the main experiments.

\pagebreak
\opdLong
{Comparison of scoring rules on the same target-generated trajectories. AUROC and TPR at 1\% FPR are averaged over fifteen targets. All scores using trained reference models use the same eight models.}
{tab:opd_scoring_rules}
{@{\extracolsep{\fill}}llcc@{}}
{Reference & Scoring rule & Mean AUROC & \makecell{Mean TPR\\at 1\% FPR}\\\midrule}
{None & Target log-probability mean & 0.529 & 0.009\\
None & Min-K\% (20\%) & 0.530 & 0.009\\
None & Min-K\%++ (20\%) & 0.511 & 0.013\\
\midrule
Base model & Mean gap & 0.520 & 0.014\\
Reference mean & Mean gap & 0.801 & 0.282\\
Reference maximum & Mean gap & 0.658 & 0.068\\
\midrule
Reference mean & ReLU & 0.774 & 0.293\\
Reference mean & \method{} & 0.777 & 0.294\\
\midrule
Reference maximum & Min-20\% gap & 0.481 & 0.008\\
Reference maximum & ReLU & 0.843 & 0.431\\
Reference maximum & \textbf{\method{}} & \textbf{0.875} & \textbf{0.465}\\}
{4}

Averaging the target's log-probabilities or selecting low-scoring tokens provides limited discrimination on the sampled trajectories. Reference-model calibration substantially improves performance, and the calibration and aggregation choices interact. Under the signed mean, the reference mean is the better reference (0.801 against 0.658 for the maximum, higher on all fifteen targets), and rectification does not help it: ReLU and \method{} lower its mean AUROC by 0.027 and 0.025 and improve only three of the fifteen targets. Under the reference maximum, ReLU and \method{} raise mean AUROC from 0.658 to 0.843 and 0.875 on all fifteen targets, and with either rectifier the maximum exceeds the mean on all fifteen targets, by 0.069 and 0.099 on average. Neither the maximum nor the rectifier contributes on its own; the gain of \method{} over the best signed-mean rule comes from their combination. This is consistent with the model of Appendix~\ref{app:leaky-negative-correction}: for a target trained without the candidate, the expected gap to the reference mean is zero for any disagreement scale $\sigma$, so both signs of the gap carry information and discarding the negative part loses signal, whereas the expected gap to the maximum is $-\sigma(m_K-m_1)$, so its signed average tracks how much the references disagree at each record and position rather than membership. Rectification removes this shortfall term and keeps the positions where the target exceeds every reference, and the small negative weight corrects the remaining chance lead.

\Needspace{5\baselineskip}
\subsection{Trajectory Generation}\label{app:opd_mia_generator_pairing}
\method{} evaluates membership on sampled trajectories, so its effectiveness may depend on how these trajectories are generated. We examine two design choices: (1) whether informative trajectories need to be generated by the training teacher, or can also be obtained from the target and other models; and (2) whether providing the reference answer during generation improves membership inference over sampling from the problem alone. These comparisons evaluate our choice to sample from the target conditioned on both the problem and reference answer. We keep the target, eight reference models, candidate records, and scoring rule fixed, and sample eight trajectories per candidate under each setting.

\noindent\textbf{Sampling model.}
For the mathematics Qwen3-1.7B target, Figure~\ref{fig:opd_ablation}(c) compares trajectories generated by the training teacher (Qwen3-8B), two alternative models (Qwen3-14B and Qwen3-4B), the frozen base model, and the target. All five models receive the problem and reference answer. The teacher, the two alternative models, and the target yield similar AUROCs, ranging from 0.899 to 0.916, while the base model yields 0.857. On this target, effective attack trajectories can thus be obtained from several models, including the target itself. Sampling from the target directly examines the generation behavior shaped by OPD.

\noindent\textbf{Reference-answer conditioning.}\label{app:leaky-reference-conditioning}
We then keep the target as the sampling model and compare trajectories generated from the problem alone with those generated from both the problem and reference answer. In both cases, the target and reference models score the trajectories using only the problem and the trajectory prefix as context. Providing the reference answer during generation increases AUROC from 0.880 to 0.916. Reference-answer conditioning therefore improves the attack on this target, while the problem-only result shows that membership information remains detectable without it.

\Needspace{5\baselineskip}
\subsection{Number of Reference Models}\label{app:opd_mia_shadow_ablation}
We vary the number of reference models to examine how much calibration can be obtained with a smaller ensemble. For each target in Table~\ref{tab:opd_main}, we keep its candidates, eight target-generated trajectories, and scoring procedure fixed, including $\alpha=0.04$. For each $K=1,\ldots,8$, we use the first $K$ models from the same set of eight reference models. The curves therefore describe a fixed sequence of reference-model subsets. Figure~\ref{fig:opd_ablation}(a,b) presents Qwen3-1.7B; Figures~\ref{fig:opd_shadow_sweep} and~\ref{fig:opd_shadow_tpr} below show the other four students across the three domains.

Even a single reference model provides membership information, with AUROC ranging from 0.612 to 0.910 across the fifteen targets. Increasing $K$ from one to eight raises AUROC by 0.056--0.251. Four reference models recover more than half of this gain on fourteen targets, and eight reference models achieve higher AUROC than four on every target. These results show that a small ensemble already provides useful calibration, while additional reference models further improve overall discrimination. TPR at 1\% FPR is also higher with eight reference models than with one on every target, but fluctuates more across intermediate values of $K$. Thus, an additional reference model does not necessarily improve detection at this particular FPR.

\begin{figure}[!htbp]\centering% BEGIN INLINE FIGURE: shadow_sweep
\definecolor{abBlue}{HTML}{74A9C5}\definecolor{abRose}{HTML}{DD7389}\definecolor{abSand}{HTML}{C4AC5E}%
\pgfplotsset{shadow panel/.style={scale only axis,width=2.8cm,height=3.2cm,axis lines*=left,axis line style={black!40,line width=.4pt},tick style={black!40,line width=.4pt},tick align=outside,major tick length=2pt,tick label style={font=\scriptsize,text=black!70},label style={font=\scriptsize,text=black!85},grid style={black!10,line width=.3pt},scaled ticks=false,clip=false,xlabel={Reference models $K$},xmin=.6,xmax=8.4,xtick={1,...,8},ymin=.6,ymax=.95,ytick={.6,.7,.8,.9},ymajorgrids,y tick label style={/pgf/number format/fixed,/pgf/number format/precision=1},title style={font=\small,yshift=-2pt,text depth=0pt}},shadow series/.style={color=#1,line width=.9pt,mark=*,mark size=1.6pt,mark options={fill=#1,draw=white,line width=.4pt}}}%
\begin{tikzpicture}
\begin{axis}[shadow panel,name=p1,title={Qwen3-4B},ylabel={AUROC},legend pos=south east,legend style={font=\scriptsize,draw=none,fill=none,inner sep=1pt,row sep=-2.5pt,cells={anchor=west}},legend image code/.code={\draw[mark repeat=2,mark phase=2,#1] plot coordinates {(0cm,0cm) (.2cm,0cm) (.4cm,0cm)};}]
\addplot[shadow series=abBlue] coordinates {(1,.7202) (2,.7949) (3,.8295) (4,.8597) (5,.8842) (6,.9046) (7,.9154) (8,.9224)};\addlegendentry{Mathematics}
\addplot[shadow series=abRose,mark=square*,mark size=1.5pt] coordinates {(1,.7654) (2,.7861) (3,.7882) (4,.7986) (5,.8208) (6,.8389) (7,.8379) (8,.8482)};\addlegendentry{Medical}
\addplot[shadow series=abSand,mark=triangle*,mark size=2pt] coordinates {(1,.6700) (2,.6876) (3,.7839) (4,.8169) (5,.8284) (6,.8404) (7,.8421) (8,.8434)};\addlegendentry{Coding}
\end{axis}
\begin{axis}[shadow panel,name=p2,at={(p1.south east)},anchor=south west,xshift=.35cm,title={Qwen3.5-2B},yticklabels={}]
\addplot[shadow series=abBlue] coordinates {(1,.7826) (2,.8365) (3,.8754) (4,.8893) (5,.8971) (6,.9040) (7,.9079) (8,.9129)};
\addplot[shadow series=abRose,mark=square*,mark size=1.5pt] coordinates {(1,.7804) (2,.8246) (3,.8454) (4,.8523) (5,.8619) (6,.8650) (7,.8682) (8,.8676)};
\addplot[shadow series=abSand,mark=triangle*,mark size=2pt] coordinates {(1,.6502) (2,.7586) (3,.8207) (4,.8497) (5,.8756) (6,.8875) (7,.8991) (8,.9013)};
\end{axis}
\begin{axis}[shadow panel,name=p3,at={(p2.south east)},anchor=south west,xshift=.35cm,title={Qwen3.5-4B},yticklabels={}]
\addplot[shadow series=abBlue] coordinates {(1,.8230) (2,.8919) (3,.9024) (4,.9063) (5,.9130) (6,.9254) (7,.9213) (8,.9251)};
\addplot[shadow series=abRose,mark=square*,mark size=1.5pt] coordinates {(1,.7342) (2,.7772) (3,.7960) (4,.8190) (5,.8275) (6,.8310) (7,.8341) (8,.8370)};
\addplot[shadow series=abSand,mark=triangle*,mark size=2pt] coordinates {(1,.6738) (2,.7424) (3,.7860) (4,.8054) (5,.8198) (6,.8311) (7,.8399) (8,.8460)};
\end{axis}
\begin{axis}[shadow panel,name=p4,at={(p3.south east)},anchor=south west,xshift=.35cm,title={Gemma-4-E2B},yticklabels={}]
\addplot[shadow series=abBlue] coordinates {(1,.6124) (2,.6732) (3,.7308) (4,.7690) (5,.8005) (6,.8148) (7,.8231) (8,.8320)};
\addplot[shadow series=abRose,mark=square*,mark size=1.5pt] coordinates {(1,.7083) (2,.7648) (3,.7902) (4,.8030) (5,.8110) (6,.8216) (7,.8281) (8,.8257)};
\addplot[shadow series=abSand,mark=triangle*,mark size=2pt] coordinates {(1,.6410) (2,.7023) (3,.7508) (4,.7792) (5,.8041) (6,.8197) (7,.8318) (8,.8393)};
\end{axis}
\end{tikzpicture}
% END INLINE FIGURE: shadow_sweep
\caption{AUROC against the number of reference models $K$ for the four students not in Figure~\ref{fig:opd_ablation}(a). Each $K$ uses the first $K$ of the same eight reference models, so $K=8$ is Table~\ref{tab:opd_main}.}\label{fig:opd_shadow_sweep}\end{figure}
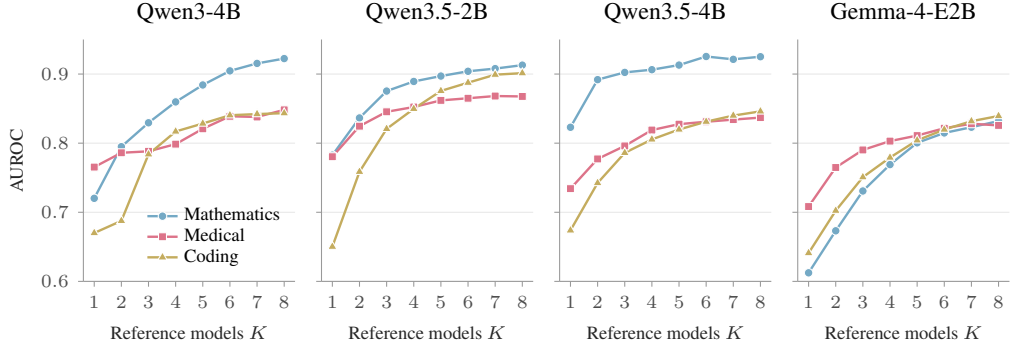
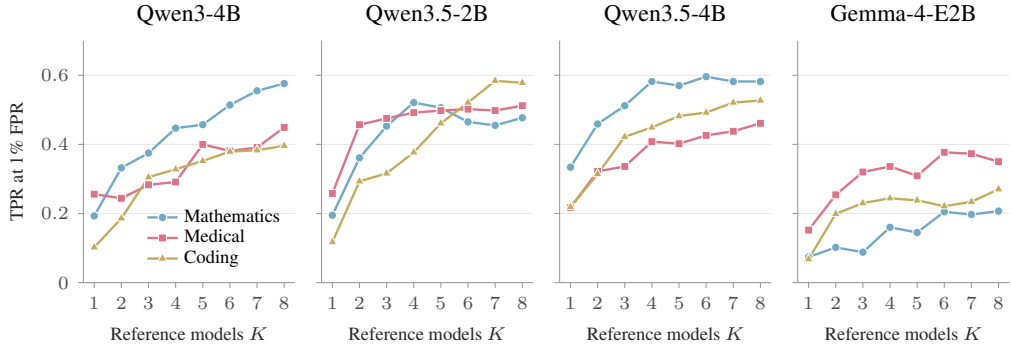
\begin{figure}[!htbp]\centering% BEGIN INLINE FIGURE: shadow_tpr
\definecolor{abBlue}{HTML}{74A9C5}\definecolor{abRose}{HTML}{DD7389}\definecolor{abSand}{HTML}{C4AC5E}%
\pgfplotsset{shadow panel/.style={scale only axis,width=2.8cm,height=3.2cm,axis lines*=left,axis line style={black!40,line width=.4pt},tick style={black!40,line width=.4pt},tick align=outside,major tick length=2pt,tick label style={font=\scriptsize,text=black!70},label style={font=\scriptsize,text=black!85},grid style={black!10,line width=.3pt},scaled ticks=false,clip=false,xlabel={Reference models $K$},xmin=.6,xmax=8.4,xtick={1,...,8},ymin=0,ymax=.7,ytick={0,.2,.4,.6},ymajorgrids,y tick label style={/pgf/number format/fixed,/pgf/number format/precision=1},title style={font=\small,yshift=-2pt,text depth=0pt}},shadow series/.style={color=#1,line width=.9pt,mark=*,mark size=1.6pt,mark options={fill=#1,draw=white,line width=.4pt}}}%
\begin{tikzpicture}
\begin{axis}[shadow panel,name=p1,title={Qwen3-4B},ylabel={TPR at 1\% FPR},legend pos=south east,legend style={font=\scriptsize,draw=none,fill=none,inner sep=1pt,row sep=-2.5pt,cells={anchor=west}},legend image code/.code={\draw[mark repeat=2,mark phase=2,#1] plot coordinates {(0cm,0cm) (.2cm,0cm) (.4cm,0cm)};}]
\addplot[shadow series=abBlue] coordinates {(1,.193) (2,.332) (3,.375) (4,.447) (5,.457) (6,.514) (7,.555) (8,.576)};\addlegendentry{Mathematics}
\addplot[shadow series=abRose,mark=square*,mark size=1.5pt] coordinates {(1,.256) (2,.244) (3,.283) (4,.291) (5,.400) (6,.381) (7,.391) (8,.449)};\addlegendentry{Medical}
\addplot[shadow series=abSand,mark=triangle*,mark size=2pt] coordinates {(1,.102) (2,.186) (3,.305) (4,.328) (5,.352) (6,.379) (7,.383) (8,.396)};\addlegendentry{Coding}
\end{axis}
\begin{axis}[shadow panel,name=p2,at={(p1.south east)},anchor=south west,xshift=.35cm,title={Qwen3.5-2B},yticklabels={}]
\addplot[shadow series=abBlue] coordinates {(1,.195) (2,.361) (3,.453) (4,.521) (5,.506) (6,.465) (7,.455) (8,.477)};
\addplot[shadow series=abRose,mark=square*,mark size=1.5pt] coordinates {(1,.258) (2,.457) (3,.475) (4,.492) (5,.498) (6,.502) (7,.498) (8,.512)};
\addplot[shadow series=abSand,mark=triangle*,mark size=2pt] coordinates {(1,.117) (2,.293) (3,.316) (4,.377) (5,.461) (6,.521) (7,.584) (8,.578)};
\end{axis}
\begin{axis}[shadow panel,name=p3,at={(p2.south east)},anchor=south west,xshift=.35cm,title={Qwen3.5-4B},yticklabels={}]
\addplot[shadow series=abBlue] coordinates {(1,.334) (2,.459) (3,.512) (4,.582) (5,.570) (6,.596) (7,.582) (8,.582)};
\addplot[shadow series=abRose,mark=square*,mark size=1.5pt] coordinates {(1,.217) (2,.322) (3,.336) (4,.408) (5,.402) (6,.426) (7,.438) (8,.461)};
\addplot[shadow series=abSand,mark=triangle*,mark size=2pt] coordinates {(1,.219) (2,.314) (3,.422) (4,.449) (5,.482) (6,.492) (7,.521) (8,.527)};
\end{axis}
\begin{axis}[shadow panel,name=p4,at={(p3.south east)},anchor=south west,xshift=.35cm,title={Gemma-4-E2B},yticklabels={}]
\addplot[shadow series=abBlue] coordinates {(1,.074) (2,.102) (3,.088) (4,.160) (5,.145) (6,.205) (7,.197) (8,.207)};
\addplot[shadow series=abRose,mark=square*,mark size=1.5pt] coordinates {(1,.152) (2,.254) (3,.320) (4,.336) (5,.309) (6,.377) (7,.373) (8,.350)};
\addplot[shadow series=abSand,mark=triangle*,mark size=2pt] coordinates {(1,.068) (2,.199) (3,.230) (4,.244) (5,.238) (6,.221) (7,.234) (8,.270)};
\end{axis}
\end{tikzpicture}
% END INLINE FIGURE: shadow_tpr
\caption{TPR at 1\% FPR along the sweeps of Figure~\ref{fig:opd_shadow_sweep}.}\label{fig:opd_shadow_tpr}\end{figure}

\Needspace{5\baselineskip}
\subsection{On-Policy Self-Distillation}\label{app:opd_mia_opsd}
We next examine whether membership leakage persists when the teacher is a frozen copy of the initial student. In on-policy self-distillation (OPSD)~\citep{zhao2026opsd,shenfeld2026sdft}, this teacher receives the problem and reference answer, while the student samples training trajectories from the problem alone. We compare OPSD with external-teacher OPD using Qwen3-1.7B on OpenThoughts-Math. The OPD arm uses Qwen3-8B as teacher and is a separate run of the main mathematics configuration.

The two runs are trained on different samples of 512 OpenThoughts-Math records, so each run has its own candidate set of 512 members and 512 non-members and its own eight reference models trained with the corresponding procedure; the runs are compared at matched training steps on cohorts of equal size. We evaluate checkpoints at 25, 50, 75, and 100 training steps. At each checkpoint, the target generates eight trajectories per candidate, and \method{} uses the reference models from the same training step with $\alpha=0.04$.

\opdLong
{Membership inference under on-policy self-distillation and external-teacher OPD on mathematics Qwen3-1.7B. Leakage increases across the evaluated checkpoints under both training procedures. TPR is measured at 1\% FPR.}
{tab:opd_opsd}
{@{\extracolsep{\fill}}l*{4}{c}@{}}
{Training step & \makecell[c]{OPSD\\AUROC} & \makecell[c]{OPSD\\TPR} & \makecell[c]{OPD\\AUROC} & \makecell[c]{OPD\\TPR}\\\midrule}
{25 & 0.5935 & 0.090 & 0.7259 & 0.260\\
50 & 0.6565 & 0.090 & 0.8088 & 0.336\\
75 & 0.6869 & 0.146 & 0.8654 & 0.424\\
100 & 0.6987 & 0.150 & 0.8771 & 0.449\\}
{5}

As shown in Table~\ref{tab:opd_opsd}, AUROC increases with training under both procedures, and the OPSD run has lower AUROC than the OPD run at every evaluated checkpoint. Its AUROC nonetheless rises from 0.594 to 0.699 (OPD: from 0.726 to 0.877), showing that self-distillation still leaves detectable membership information. A control that replaces the target with a reference model yields AUROCs of 0.490--0.533 across the checkpoints of both runs. Together, these observations show that membership leakage is also present in the evaluated self-distilled student as it learns from reference-conditioned supervision.

\subsection{Blackbox setting}\label{app:opd_mia_blackbox}
We examine whether the target's predictions still reveal membership when its token probabilities are unavailable. The interface accepts a problem and trajectory prefix and returns one greedy next token. The attacker first samples trajectories from the target conditioned on the problem and reference answer, and then compares the target's returned tokens with the predictions of its own reference models.

\noindent\textbf{Black-box score.}
Let $h_{i,t}=(x,y_{i,<t})$ be the prefix preceding a sampled token $y_{i,t}$. Define $E_{i,t}=1$ when the target's greedy prediction at this prefix equals $y_{i,t}$, and zero otherwise. Similarly, let $G_{i,t}=1$ when none of the reference models predicts $y_{i,t}$, and zero otherwise. We compute
\begin{equation}
S_{\mathrm{BB}}(z)=
\frac{\sum_i\sum_{t=1}^{L_i}\left[G_{i,t}E_{i,t}-\alpha(1-G_{i,t})(1-E_{i,t})\right]}{\sum_i L_i},
\qquad \alpha=0.04.
\label{eq:opd_blackbox_score}
\end{equation}
The positive term counts positions where only the target predicts the sampled token. The negative term counts positions where at least one reference model predicts it but the target does not. This preserves the asymmetric treatment of target advantages and shortfalls using discrete predictions. The score pools all scored positions across trajectories, with every position contributing to the denominator.

\noindent\textbf{Evaluation and additional results.}
We use the same fifteen targets and candidate sets as the main experiments, with eight target-generated trajectories and eight reference models. For scoring, the target and reference models receive the problem and trajectory prefix. We compute the score from greedy decisions at temperature zero, using only the identity of the predicted token. Implementation details are given in Appendix~\ref{app:opd_mia_primary_protocol}. Table~\ref{tab:opd_blackbox_full} supplements the main black-box results with TPR at 5\% and 10\% FPR. Replacing the target with one reference model and using the remaining seven as references gives AUROCs of 0.474--0.541.

\opdLong
{Additional black-box results at 5\% and 10\% FPR for the fifteen targets in Table~\ref{tab:opd_mia_blackbox}.}
{tab:opd_blackbox_full}
{@{\extracolsep{\fill}}lcc@{}}
{Student & \makecell{TPR at\\5\% FPR} & \makecell{TPR at\\10\% FPR}\\\midrule}
{\addlinespace[3pt]\multicolumn{3}{@{}l}{\textit{Mathematics}}\\*[2pt]
Qwen3-1.7B & 0.393 & 0.533\\
Qwen3-4B & 0.453 & 0.543\\
Qwen3.5-2B & 0.344 & 0.486\\
Qwen3.5-4B & 0.469 & 0.568\\
Gemma-4-E2B & 0.283 & 0.396\\
\addlinespace[3pt]\multicolumn{3}{@{}l}{\textit{Medical question answering}}\\*[2pt]
Qwen3-1.7B & 0.590 & 0.648\\
Qwen3-4B & 0.283 & 0.350\\
Qwen3.5-2B & 0.453 & 0.523\\
Qwen3.5-4B & 0.324 & 0.439\\
Gemma-4-E2B & 0.293 & 0.365\\
\addlinespace[3pt]\multicolumn{3}{@{}l}{\textit{Code generation}}\\*[2pt]
Qwen3-1.7B & 0.352 & 0.469\\
Qwen3-4B & 0.250 & 0.322\\
Qwen3.5-2B & 0.352 & 0.410\\
Qwen3.5-4B & 0.488 & 0.557\\
Gemma-4-E2B & 0.221 & 0.311\\}
{3}

\begin{figure}[!htbp]\centering% BEGIN INLINE FIGURE: bb_continuations
\definecolor{abBlue}{HTML}{74A9C5}\definecolor{abRose}{HTML}{DD7389}\definecolor{abSand}{HTML}{C4AC5E}%
\pgfplotsset{sm panel/.style={scale only axis,axis lines*=left,axis line style={black!40,line width=.4pt},tick style={black!40,line width=.4pt},tick align=outside,major tick length=2pt,tick label style={font=\scriptsize,text=black!70},label style={font=\scriptsize,text=black!85},xlabel style={text depth=.3ex},ylabel style={align=center},grid style={black!10,line width=.3pt},scaled ticks=false,clip=false,title style={font=\small,yshift=-2pt,text depth=0pt},legend style={font=\scriptsize,draw=none,fill=none,inner sep=1pt,cells={anchor=west},/tikz/every even column/.append style={column sep=7pt}},legend image code/.code={\draw[mark repeat=2,mark phase=2,##1] plot coordinates {(0cm,0cm) (.2cm,0cm) (.4cm,0cm)};}},sm series/.style={color=#1,line width=.9pt,mark=*,mark size=1.6pt,mark options={fill=#1,draw=white,line width=.4pt}},dom math/.style={sm series=abBlue},dom med/.style={sm series=abRose,mark=square*,mark size=1.5pt},dom code/.style={sm series=abSand,mark=triangle*,mark size=2pt},sm dot/.style={only marks,mark options={fill=#1,draw=white,line width=.4pt}},dot math/.style={sm dot=abBlue,mark=*,mark size=1.8pt},dot med/.style={sm dot=abRose,mark=square*,mark size=1.6pt},dot code/.style={sm dot=abSand,mark=triangle*,mark size=2.2pt},fixed2/.style={/pgf/number format/fixed,/pgf/number format/precision=2},fixed1/.style={/pgf/number format/fixed,/pgf/number format/precision=1}}%
\begin{tikzpicture}
\begin{axis}[sm panel,name=r0c0,width=2.25cm,height=2.1cm,xmode=log,log basis x={2},xmin=.8,xmax=10,xtick={1,2,4,8},xminorticks=false,ymajorgrids,y tick label style={fixed1},at={(0,0)},anchor=north west,xticklabels={},ymin=.55,ymax=.9,ytick={.6,.7,.8,.9},title={Qwen3-1.7B},ylabel={AUROC}]
\addplot[dom math,forget plot] coordinates {(1,0.6388) (2,0.6786) (4,0.7207) (8,0.7764)};
\addplot[dom med,forget plot] coordinates {(1,0.6908) (2,0.7411) (4,0.8115) (8,0.8545)};
\addplot[dom code,forget plot] coordinates {(1,0.6079) (2,0.6714) (4,0.7057) (8,0.7465)};
\end{axis}
\begin{axis}[sm panel,name=r0c1,width=2.25cm,height=2.1cm,xmode=log,log basis x={2},xmin=.8,xmax=10,xtick={1,2,4,8},xminorticks=false,ymajorgrids,y tick label style={fixed1},at={(r0c0.south east)},anchor=south west,xshift=.3cm,xticklabels={},ymin=.55,ymax=.9,ytick={.6,.7,.8,.9},title={Qwen3-4B},yticklabels={}]
\addplot[dom math,forget plot] coordinates {(1,0.6671) (2,0.7030) (4,0.7415) (8,0.7958)};
\addplot[dom med,forget plot] coordinates {(1,0.6119) (2,0.6456) (4,0.6821) (8,0.6908)};
\addplot[dom code,forget plot] coordinates {(1,0.5952) (2,0.6189) (4,0.6664) (8,0.6871)};
\end{axis}
\begin{axis}[sm panel,name=r0c2,width=2.25cm,height=2.1cm,xmode=log,log basis x={2},xmin=.8,xmax=10,xtick={1,2,4,8},xminorticks=false,ymajorgrids,y tick label style={fixed1},at={(r0c1.south east)},anchor=south west,xshift=.3cm,xticklabels={},ymin=.55,ymax=.9,ytick={.6,.7,.8,.9},title={Qwen3.5-2B},yticklabels={}]
\addplot[dom math,forget plot] coordinates {(1,0.6107) (2,0.6638) (4,0.7163) (8,0.7738)};
\addplot[dom med,forget plot] coordinates {(1,0.6349) (2,0.7030) (4,0.7369) (8,0.7677)};
\addplot[dom code,forget plot] coordinates {(1,0.5856) (2,0.6405) (4,0.7039) (8,0.7403)};
\end{axis}
\begin{axis}[sm panel,name=r0c3,width=2.25cm,height=2.1cm,xmode=log,log basis x={2},xmin=.8,xmax=10,xtick={1,2,4,8},xminorticks=false,ymajorgrids,y tick label style={fixed1},at={(r0c2.south east)},anchor=south west,xshift=.3cm,xticklabels={},ymin=.55,ymax=.9,ytick={.6,.7,.8,.9},title={Qwen3.5-4B},yticklabels={}]
\addplot[dom math,forget plot] coordinates {(1,0.6442) (2,0.7212) (4,0.7723) (8,0.8117)};
\addplot[dom med,forget plot] coordinates {(1,0.6051) (2,0.6436) (4,0.6993) (8,0.7371)};
\addplot[dom code,forget plot] coordinates {(1,0.6675) (2,0.7134) (4,0.7581) (8,0.7877)};
\end{axis}
\begin{axis}[sm panel,name=r0c4,width=2.25cm,height=2.1cm,xmode=log,log basis x={2},xmin=.8,xmax=10,xtick={1,2,4,8},xminorticks=false,ymajorgrids,y tick label style={fixed1},at={(r0c3.south east)},anchor=south west,xshift=.3cm,xticklabels={},ymin=.55,ymax=.9,ytick={.6,.7,.8,.9},title={Gemma-4-E2B},yticklabels={}]
\addplot[dom math,forget plot] coordinates {(1,0.6158) (2,0.6121) (4,0.6642) (8,0.7274)};
\addplot[dom med,forget plot] coordinates {(1,0.6062) (2,0.6348) (4,0.6608) (8,0.6981)};
\addplot[dom code,forget plot] coordinates {(1,0.5828) (2,0.5806) (4,0.6169) (8,0.6599)};
\end{axis}
\begin{axis}[sm panel,name=r1c0,width=2.25cm,height=2.1cm,xmode=log,log basis x={2},xmin=.8,xmax=10,xtick={1,2,4,8},xminorticks=false,ymajorgrids,y tick label style={fixed1},at={(r0c0.south west)},anchor=north west,yshift=-0.35cm,xticklabels={1,2,4,8},xlabel={Trajectories $N$},ymin=0,ymax=.45,ytick={0,.2,.4},ylabel={TPR at 1\% FPR}]
\addplot[dom math,forget plot] coordinates {(1,0.098) (2,0.152) (4,0.209) (8,0.273)};
\addplot[dom med,forget plot] coordinates {(1,0.172) (2,0.211) (4,0.305) (8,0.389)};
\addplot[dom code,forget plot] coordinates {(1,0.104) (2,0.158) (4,0.156) (8,0.213)};
\end{axis}
\begin{axis}[sm panel,name=r1c1,width=2.25cm,height=2.1cm,xmode=log,log basis x={2},xmin=.8,xmax=10,xtick={1,2,4,8},xminorticks=false,ymajorgrids,y tick label style={fixed1},at={(r1c0.south east)},anchor=south west,xshift=.3cm,xticklabels={1,2,4,8},xlabel={Trajectories $N$},ymin=0,ymax=.45,ytick={0,.2,.4},yticklabels={}]
\addplot[dom math,forget plot] coordinates {(1,0.125) (2,0.176) (4,0.273) (8,0.289)};
\addplot[dom med,forget plot] coordinates {(1,0.066) (2,0.117) (4,0.135) (8,0.211)};
\addplot[dom code,forget plot] coordinates {(1,0.062) (2,0.086) (4,0.102) (8,0.146)};
\end{axis}
\begin{axis}[sm panel,name=r1c2,width=2.25cm,height=2.1cm,xmode=log,log basis x={2},xmin=.8,xmax=10,xtick={1,2,4,8},xminorticks=false,ymajorgrids,y tick label style={fixed1},at={(r1c1.south east)},anchor=south west,xshift=.3cm,xticklabels={1,2,4,8},xlabel={Trajectories $N$},ymin=0,ymax=.45,ytick={0,.2,.4},yticklabels={},legend style={at={(.5,-.62)},anchor=north,legend columns=-1}]
\addplot[dom math,forget plot] coordinates {(1,0.068) (2,0.074) (4,0.141) (8,0.191)};
\addplot[dom med,forget plot] coordinates {(1,0.127) (2,0.203) (4,0.256) (8,0.342)};
\addplot[dom code,forget plot] coordinates {(1,0.070) (2,0.119) (4,0.168) (8,0.238)};
\addlegendimage{dom math}\addlegendentry{Mathematics}
\addlegendimage{dom med}\addlegendentry{Medical}
\addlegendimage{dom code}\addlegendentry{Coding}
\end{axis}
\begin{axis}[sm panel,name=r1c3,width=2.25cm,height=2.1cm,xmode=log,log basis x={2},xmin=.8,xmax=10,xtick={1,2,4,8},xminorticks=false,ymajorgrids,y tick label style={fixed1},at={(r1c2.south east)},anchor=south west,xshift=.3cm,xticklabels={1,2,4,8},xlabel={Trajectories $N$},ymin=0,ymax=.45,ytick={0,.2,.4},yticklabels={}]
\addplot[dom math,forget plot] coordinates {(1,0.148) (2,0.170) (4,0.277) (8,0.309)};
\addplot[dom med,forget plot] coordinates {(1,0.094) (2,0.119) (4,0.166) (8,0.189)};
\addplot[dom code,forget plot] coordinates {(1,0.131) (2,0.172) (4,0.270) (8,0.412)};
\end{axis}
\begin{axis}[sm panel,name=r1c4,width=2.25cm,height=2.1cm,xmode=log,log basis x={2},xmin=.8,xmax=10,xtick={1,2,4,8},xminorticks=false,ymajorgrids,y tick label style={fixed1},at={(r1c3.south east)},anchor=south west,xshift=.3cm,xticklabels={1,2,4,8},xlabel={Trajectories $N$},ymin=0,ymax=.45,ytick={0,.2,.4},yticklabels={}]
\addplot[dom math,forget plot] coordinates {(1,0.033) (2,0.086) (4,0.119) (8,0.152)};
\addplot[dom med,forget plot] coordinates {(1,0.080) (2,0.111) (4,0.145) (8,0.189)};
\addplot[dom code,forget plot] coordinates {(1,0.039) (2,0.033) (4,0.055) (8,0.084)};
\end{axis}
\end{tikzpicture}
% END INLINE FIGURE: bb_continuations
\caption{Effect of the number of target-generated trajectories per candidate across all fifteen targets. Eight reference models and $\alpha=0.04$ are fixed throughout. The eight-trajectory results match Table~\ref{tab:opd_mia_blackbox}.}\label{fig:opd_bb_continuations}\end{figure}
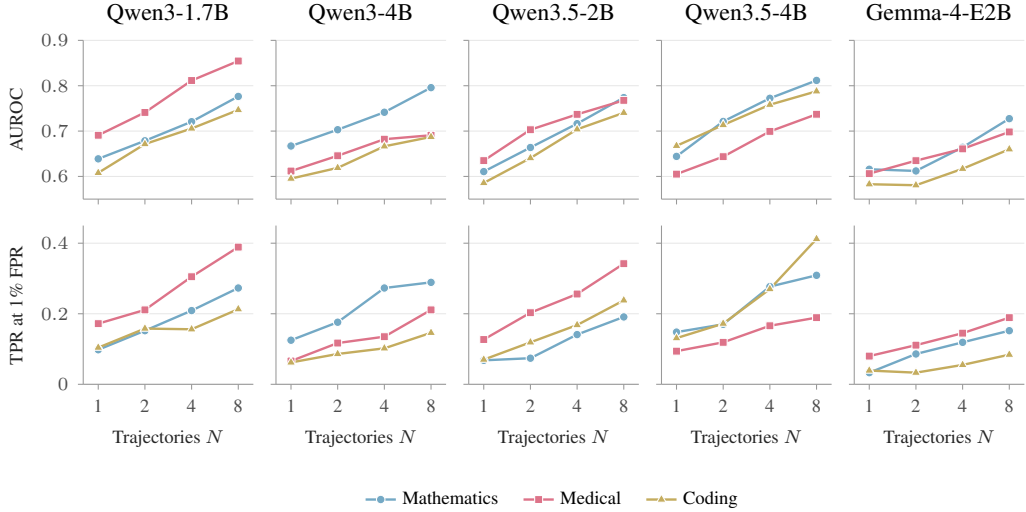

\Needspace{4\baselineskip}
\noindent\textbf{Number of sampled trajectories.}
To examine how the number of trajectories affects detection, we use the first one, two, four, or eight trajectories from each candidate's sampled set while keeping the eight reference models and scoring rule fixed. Figure~\ref{fig:opd_bb_continuations} shows that eight trajectories yield higher AUROC and TPR than one on all fifteen targets. Additional trajectories provide more positions at which to compare target and reference predictions and identify membership signals.

\section{Defense Evaluation}\label{app:opd_mia_defense}
\subsection{Defense Setup}
We evaluate whether LoRA and DP-LoRA reduce membership leakage from OPD. The experiments use Qwen3-1.7B with a Qwen3-8B teacher across mathematics, medical question answering, and code generation. Within each domain, the training data and evaluation candidates remain fixed across full fine-tuning, LoRA, and DP-LoRA. Each configuration has its own target and eight reference models trained with the corresponding procedure. We evaluate \method{} using eight trajectories sampled from each target for every candidate.

\noindent\textbf{LoRA.} Both LoRA configurations apply rank-64 adapters ($\alpha=128$) to all linear layers of the transformer blocks. LoRA without DP uses the training budget of full fine-tuning (126 steps with a global batch size of 32 for mathematics; 512 steps with a global batch size of 128 for medical QA and coding), no gradient clipping, and the interpolation weight of the corresponding full fine-tuning cell. 

\noindent\textbf{DP-LoRA.} DP-LoRA trains the same adapters with per-record gradient clipping and Gaussian noise at a nominal privacy budget of $\varepsilon=8$ with $\delta=1/n$, where $n$ is the number of training records; the noise multiplier is computed with the R\'enyi-DP accountant~\citep{mironov2017rdp} of Opacus~\citep{yousefpour2021opacus} for each training configuration. Mathematics and coding use the schedule of full fine-tuning with clipping norm $C=1$ and a learning rate of $10^{-4}$ (noise multipliers 0.698 and 0.60). For medical QA, we report a configuration at the same privacy budget and the same number of processed records that uses $C=0.25$ and a learning rate of $3\times10^{-4}$.

\subsection{Full Results}
Table~\ref{tab:opd_defense_full} supplements Table~\ref{tab:opd_mia_defense} with TPR at 5\% and 10\% FPR.

% BEGIN INLINE TABLE: defense_full
\opdLong
{Defenses on Qwen3-1.7B: change in Avg@8 over the base (percentage points), AUROC, and TPR at 1\%, 5\%, and 10\% FPR of \method{} ($K=8$, $\alpha=0.04$).}
{tab:opd_defense_full}
{@{\extracolsep{\fill}}llrcccc@{}}
{Domain & Training & $\Delta$Util. & AUROC & TPR@1\% & TPR@5\% & TPR@10\%\\\midrule}
{Mathematics & Full fine-tuning & $+5.00$ & 0.9158 & 0.418 & 0.662 & 0.758\\
 & LoRA & $+2.50$ & 0.6455 & 0.098 & 0.180 & 0.277\\
 & DP-LoRA & $+1.67$ & 0.4903 & 0.014 & 0.049 & 0.092\\
\midrule
Medical QA & Full fine-tuning & $+3.98$ & 0.9665 & 0.740 & 0.865 & 0.928\\
 & LoRA & $+2.46$ & 0.7868 & 0.268 & 0.396 & 0.502\\
 & DP-LoRA & $+0.04$ & 0.5044 & 0.023 & 0.070 & 0.111\\
\midrule
Code generation & Full fine-tuning & $+0.57$ & 0.8467 & 0.434 & 0.568 & 0.633\\
 & LoRA & $+0.14$ & 0.6585 & 0.146 & 0.236 & 0.338\\
 & DP-LoRA & $+0.07$ & 0.5097 & 0.027 & 0.059 & 0.115\\}
{7}

\subsection{Results across Attack Configurations}
The main results show that LoRA reduces attack performance while retaining detectable membership information, and DP-LoRA brings AUROC close to chance. We examine whether this observation depends on the default \method{} configuration by changing the negative slope or the number of reference models on the three domains. Within each training configuration, the candidates and eight target-generated trajectories remain fixed. We compare the default score ($K=8$, $\alpha=0.04$) with positive-only aggregation ($\alpha=0$), and with four or one reference model while retaining $\alpha=0.04$.

\opdLong
{Attack performance under LoRA and DP-LoRA across four scoring configurations. Entries report AUROC / TPR at 1\% FPR.}
{tab:opd_defense_configurations}
{@{\extracolsep{\fill}}llcc@{}}
{Domain & Attack configuration & LoRA & DP-LoRA\\\midrule}
{Mathematics & \method{} ($K=8$, $\alpha=0.04$) & 0.6455 / 0.098 & 0.4903 / 0.014\\
 & ReLU ($K=8$, $\alpha=0$) & 0.5881 / 0.084 & 0.4874 / 0.016\\
 & \method{} ($K=4$, $\alpha=0.04$) & 0.5866 / 0.076 & 0.4856 / 0.008\\
 & \method{} ($K=1$, $\alpha=0.04$) & 0.5511 / 0.039 & 0.4903 / 0.008\\
\midrule
Medical QA & \method{} ($K=8$, $\alpha=0.04$) & 0.7868 / 0.268 & 0.5044 / 0.023\\
 & ReLU ($K=8$, $\alpha=0$) & 0.7689 / 0.240 & 0.5025 / 0.021\\
 & \method{} ($K=4$, $\alpha=0.04$) & 0.7577 / 0.229 & 0.5000 / 0.010\\
 & \method{} ($K=1$, $\alpha=0.04$) & 0.6966 / 0.141 & 0.5041 / 0.018\\
\midrule
Code generation & \method{} ($K=8$, $\alpha=0.04$) & 0.6585 / 0.146 & 0.5097 / 0.027\\
 & ReLU ($K=8$, $\alpha=0$) & 0.6309 / 0.107 & 0.5131 / 0.027\\
 & \method{} ($K=4$, $\alpha=0.04$) & 0.6292 / 0.074 & 0.5172 / 0.035\\
 & \method{} ($K=1$, $\alpha=0.04$) & 0.5653 / 0.041 & 0.5399 / 0.014\\}
{4}

Table~\ref{tab:opd_defense_configurations} shows that LoRA remains above chance across these configurations, with AUROC ranging from 0.551 to 0.787. For DP-LoRA, AUROC stays between 0.486 and 0.540. The low attack AUROC under DP-LoRA therefore persists when the negative contribution is removed or the reference ensemble is reduced.

\end{document}